\documentclass{sig-alternate} 
\usepackage{mathptmx} 

\usepackage{fancyhdr}
\usepackage[normalem]{ulem}
\usepackage{xcolor}

\usepackage{balance}
\usepackage{flushend}
\usepackage{multirow}
\usepackage{enumerate}
\usepackage{caption}
\usepackage{subcaption}
\usepackage{romannum}

\usepackage[hyphens]{url}
\usepackage[hidelinks]{hyperref}
\hypersetup{breaklinks=true}
\usepackage{cite}

\usepackage{mathtools}
\usepackage{booktabs}
\usepackage{multirow}

\usepackage{nicefrac}

\usepackage{algorithm}
\usepackage[noend]{algpseudocode}
\usepackage{colortbl}

\usepackage{pifont}
\newcommand{\cmark}{\ding{51}}%
\newcommand{\xmark}{\ding{55}}%

\newcommand{\ignore}[1]{}
\newcommand{\needcite}[1]{{\bf\color{red}[]}}

\newcommand*{\x}{\mathsf{x}\mskip1mu}

\title{Musical Chair: Efficient Real-Time Recognition \\Using Collaborative IoT Devices} 
\numberofauthors{5} 
\author{
\alignauthor
Ramyad Hadidi\\
       \affaddr{Georgia Institute of Technology}\\
\alignauthor
Jiashen Cao\\
       \affaddr{Georgia Institute of Technology}\\
\alignauthor 
Matthew Woodward\\
       \affaddr{Georgia Institute of Technology}\\
\and
\alignauthor 
Michael S. Ryoo\\
       \affaddr{Indiana University\\EgoVid Inc.}\\
\alignauthor 
Hyesoon Kim\\
       \affaddr{Georgia Institute of Technology}\\
}

\begin{document}
\maketitle
\pagestyle{plain}
\pagenumbering{arabic}

\begin{abstract}

The prevalence of Internet of things (IoT) devices and abundance of sensor data has created an increase in real-time data processing such as recognition of speech, image, and video. While currently such processes are offloaded to the computationally powerful cloud system, a localized and distributed approach is desirable because (\romannum{1}) it preserves the privacy of users and (\romannum{2}) it omits the dependency on cloud services. However, IoT networks are usually composed of resource-constrained devices, and a single device is not powerful enough to process real-time data. To overcome this challenge, we examine data and model parallelism for such devices in the context of deep neural networks. We propose Musical Chair to enable efficient, localized, and dynamic real-time recognition by harvesting the aggregated computational power from the resource-constrained devices in the same IoT network as input sensors. Musical chair adapts to the availability of computing devices at runtime and adjusts to the inherit dynamics of IoT networks. To demonstrate Musical Chair, on a network of Raspberry PIs (up to 12) each connected to a camera, we implement a state-of-the-art action recognition model for videos and two recognition models for images. Compared to the Tegra TX2, an embedded low-power platform with a six-core CPU and a GPU, our distributed action recognition system achieves not only similar energy consumption but also twice the performance of the TX2. Furthermore, in image recognition, Musical Chair achieves similar performance and saves dynamic energy.

\end{abstract}

\section{Introduction}

\noindent
The ever-increasing number of Internet of things (IoT) devices, which outnumbered the world's population this year\-~\cite{gartner-iot}, generate large quantities of raw data that need to be processed and analyzed in real time. Processing large quantities of IoT data, besides introducing new challenges for privacy and security, dramatically increases network traffic and load on the cloud (i.e., data centers)~\cite{gartner-iot-datacenter,lee:lee15,khan:khan12}. In addition, the fast-paced advancements of deep neural network (DNN) research extends capabilities of DNN for tasks that are suitable for IoT devices, such as computer vision~\cite{kri:sut12}, natural language processing~\cite{col:wes08}, neural machine translation~\cite{bah:cho14}, and video recognition~\cite{ryoo:kim17, sim:zis14}. However, since performing these tasks is often a challenge in resource-constrained IoT devices, computations of DNNs (inference) are offloaded to the cloud. Therefore, introducing such phenomenal DNN capabilities further increases the load of IoT devices on the cloud while increasing the dependency of devices on the availability of data centers with a high quality of service and network resources. As a result, a significant amount of research efforts has been invested for overcoming these challenges exacerbated by DNN applications in resource-constrained IoT devices~\cite{wan:li16, kim:par15, lei:sen13, mcd:tee17, ban:wan17, lik:hou16}, such as collaborative computation between edge devices and the cloud~\cite{kan:hau17, hau:man14, tee:mcd17}, or customized mobile implementations~\cite{ell, rast:ord16, how:zhu17, han:shn16, caffe2Go, tensorflowLite, ian:mos16, red:kum16}. Despite all these efforts, scaling current DNN applications to IoT devices and process generated data in real time still raises privacy concerns and requires significant cloud resources or increases the financial cost for the user. Hence, to handle the current and future more resource-hungry DNN applications~\cite{he:zha16, sim:zis14-deep, sze:liu15}, creating a cost-effective and efficient solution in IoT networks that does not rely on data centers or network resources is critical.

State-of-the-art IoT networks are formed with various IoT sensors and recording agents, such as HD cameras~\cite{nest} and temperature sensors~\cite{nest-term}, many of which are capable of performing small computations within their processor~\cite{nest-term-breakdown,nest-term-breakdown-arm,nest-6core}. In fact, the low cost and widespread availability of such low-power processors has accelerated the integration of IoT devices as a driving force. Nevertheless, processing sensor's data in real time using DNN models is still a challenge for IoT and embedded devices. There have been many studies to overcome this challenge while preserving the accuracy of DNN models, such as pruning~\cite{yu:luk17,han:mao15}, resource partitioning~\cite{she:fer16, guo:yin17}, quantization and low-precision inference~\cite{cou:ben14, gon:li14, van:sen11}, and binarizing weights~\cite{li:zha16, cou:hub:16,rast:ord16}. Although these methods reduce the computation overhead of DNNs, they still require noticeable hardware resources for computation~\cite{can:pas16}. To solve this challenge in resource-constrained IoT devices, we can utilize the aggregated computational power of already connected IoT devices to perform DNN-based recognition in real time. Such distribution, while scalable, reduces the dependency on cloud services and preserves the privacy of the users. 

\begin{figure}[b]
\centering
\vspace{-15pt}
\includegraphics[width=1.0\linewidth]{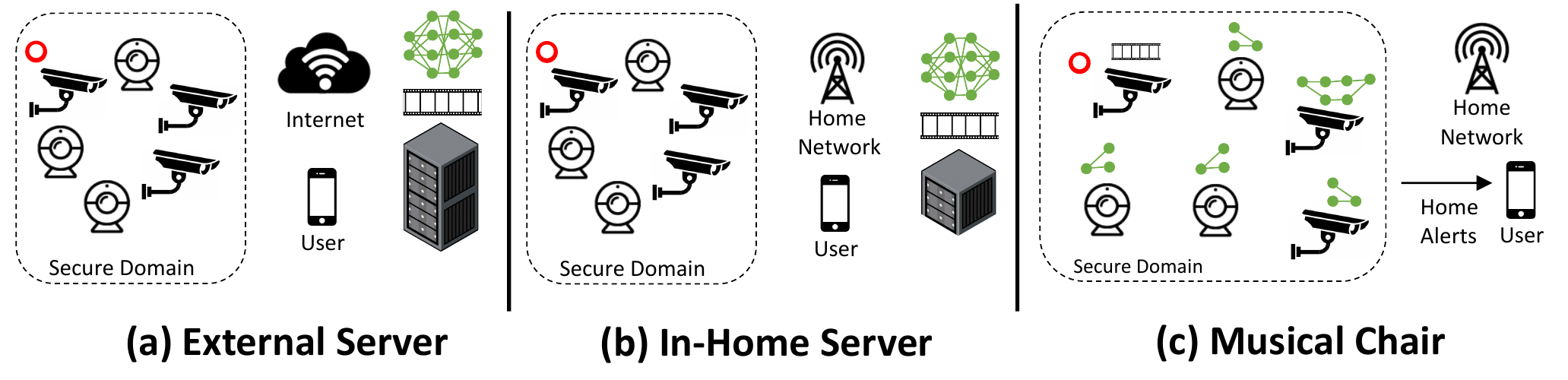}
\captionsetup{singlelinecheck=on,aboveskip=5pt, belowskip=0pt}
\caption{Three approaches for real-time DNN.}
\label{fig:intro}
\vspace{-0pt}
\end{figure}

\begin{figure}[t]
\centering
\vspace{0pt}
\includegraphics[width=1.0\linewidth]{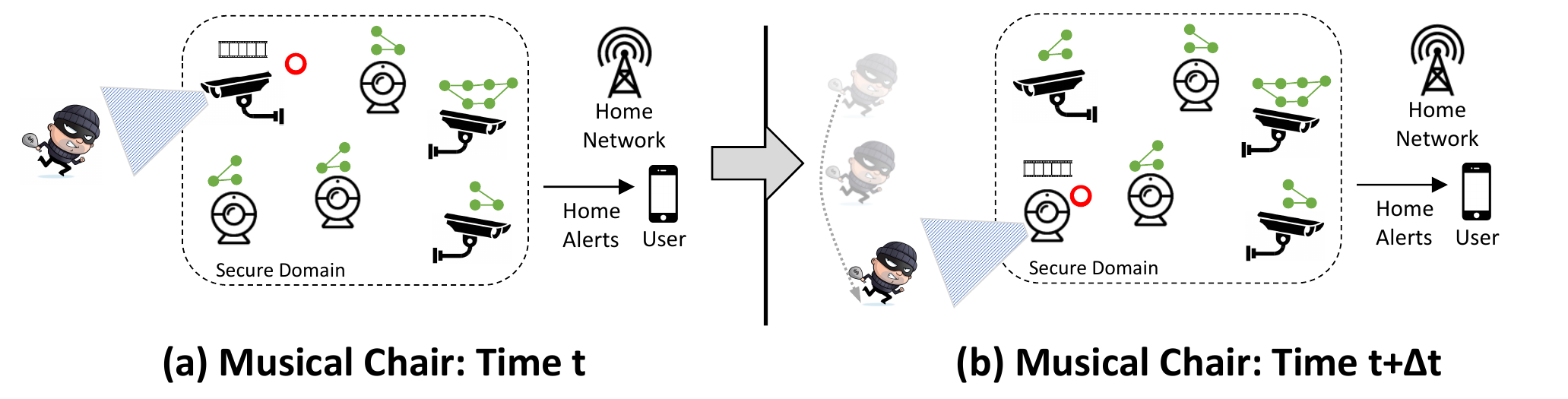}
\captionsetup{singlelinecheck=on,aboveskip=5pt, belowskip=0pt}
\caption{Musical Chair dynamically adjusts to input.}
\label{fig:intro-2}
\vspace{-15pt}
\end{figure}

In this paper, we propose Musical Chair, in which collaborative, low-power, and resource-constrained IoT devices perform cost-efficient, real-time, and dynamic DNN-based recognition (image and video). Figure~\ref{fig:intro} illustrates three different approaches for processing real-time recognition. As discussed, using an external server causes a significant increase in network traffic and computation load to the server. Another approach is to use a powerful in-home server, which is exposed to the unsecured home network and increases financial costs to the user. On the other hand, by dynamically adjusting to input conditions (Figure~\ref{fig:intro-2}), Musical Chair exploits the collective computing power of IoT devices to perform real-time recognition. To demonstrate Musical Chair, we implement an IoT network with different numbers of Raspberry PIs~\cite{pi3} (up to 12), each of which is connected to a camera~\cite{pi3-cam}. As an example for DNN, to detect an object and related type of actions happening in an environment, we thoroughly investigate a state-of-the-art action recognition DNN model~\cite{ryoo:kim17} with 15 layers. For instance, such a model can be implemented in a smart home environment for detecting threats, such as burglaries and thefts. To demonstrate our solution further, we also apply Musical Chair to well-known image recognition models and present the result.

We explore both data parallelism and model parallelism, where data parallelism consists of processing independent data concurrently for convolution layers and task parallelism consists of splitting the computation of one layer across multiple devices for fully connected layers. 
We find that because of the memory limitation in the resource-constrained devices, some tasks should utilize model parallelism while computationally bounded tasks should utilize data parallelism.
The contributions of Musical Chair are as follows:
(\romannum{1}) This is the first work that distributes a state-of-the-art action recognition model among multiple IoT devices and performs predictions solely with IoT devices. 
(\romannum{2})  This is the first work that dynamically changes the distributed tasks to interchange between sensor input devices and computational devices. 
(\romannum{3}) It proposes a complete solution for distributing DNN models among resource-constrained devices by considering model and data parallelism, memory usage, communication overhead, and real-time data processing performance.

\section{Background}
\label{sec:back}

\noindent This section overviews DNN layers, image and action recognition with DNNs, and introduces the two-stream CNN model.

\subsection{Layers in DNN}
\label{sec:ML-layers}

\noindent
This section discusses layers employed in modern DNNs.

\noindent {\bf Fully Connected Layer \texttt{(fc)}:} A fully connected (dense) layer of size $n$ has $n$ weights, and, on its input data, performs a linear operation of weighted summation that creates an output of size $n$.

\noindent {\bf Convolution Layer \texttt{(conv)}:} A CNN layer, consists of a set of filters that are applied to a subset of inputs by sweeping each filter (i.e, kernel) over them. Performed operations are linear (matrix multiplications).

\noindent {\bf Batch Normalization Layer \texttt{(norm)}:} A batch normalization layer enforces a unit Gaussian distribution (mean to zero, and variance to one) over the input data (activations) of next layer~\cite{iof:sze15}, and helps us by reducing learning time and increasing overall accuracy.

\noindent {\bf Max Pooling Layer \texttt{(maxpool)}:} A pooling layer, without significantly affecting valid data, effectively downsamples the output of prior layer; therefore, reduces the dimensionality of data, and the number of required operation in the following layers. In addition, since it outputs an abstracted version of data, it prevents overfitting. 

\noindent {\bf Activation Layer \texttt{(act)}:} An activation layer applies a non-linear function over the data allowing a model to learn complex functions. In this paper, we use rectified linear unit (ReLU), $f(x)=max(0,x)$, after \texttt{conv} layers. 

\noindent {\bf Softmax Layer  \texttt{(softmax)}:} A softmax layer is implemented as an activation layer for the final fully connected layer to perform a classification. In detail, this layer generates a categorical probability distribution for each class in the output.

\subsection{Image Recognition}
\label{sec:image-recognition}

\begin{figure}[h]
\vspace{-8pt}
\centering
\begin{tabular}{c}
\vspace{0pt}

\begin{subfigure}{\columnwidth}
\centering
\vspace{0pt}
\includegraphics[width=1.0\linewidth]{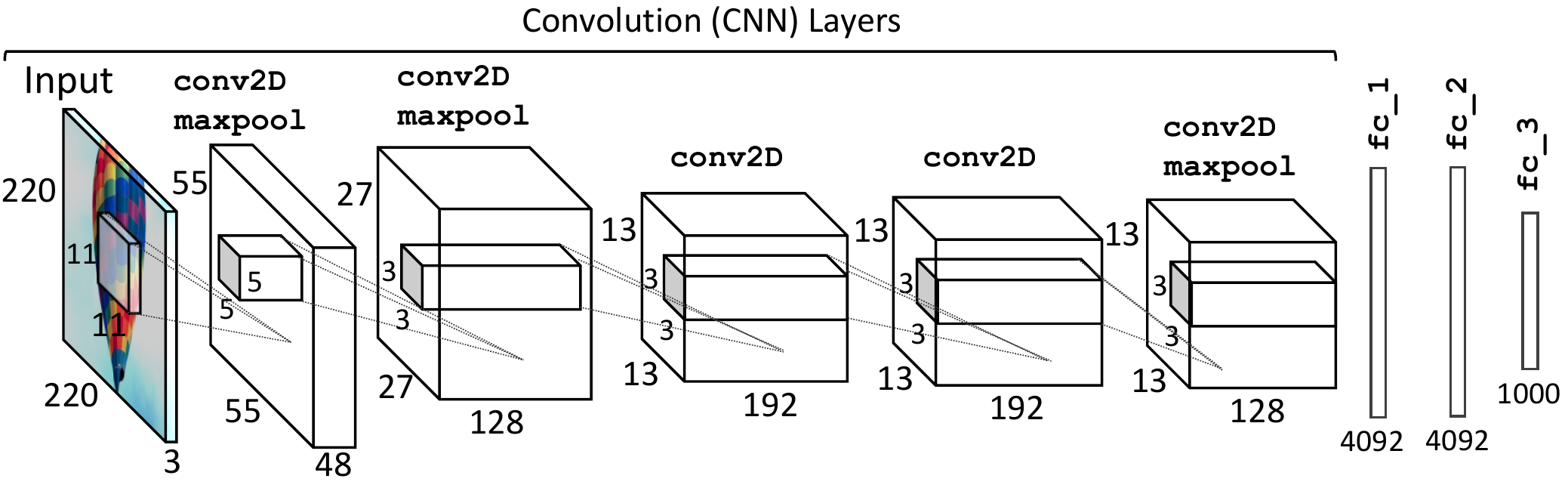}
\captionsetup{singlelinecheck=on,aboveskip=-2pt, belowskip=0pt}
\caption{Single stream AlexNet model.}
\label{fig:alexnet}
\vspace{0pt}
\end{subfigure}
\vspace{5pt}

\\

\begin{subfigure}{\columnwidth}
\centering
\vspace{0pt}
\includegraphics[width=1.0\linewidth]{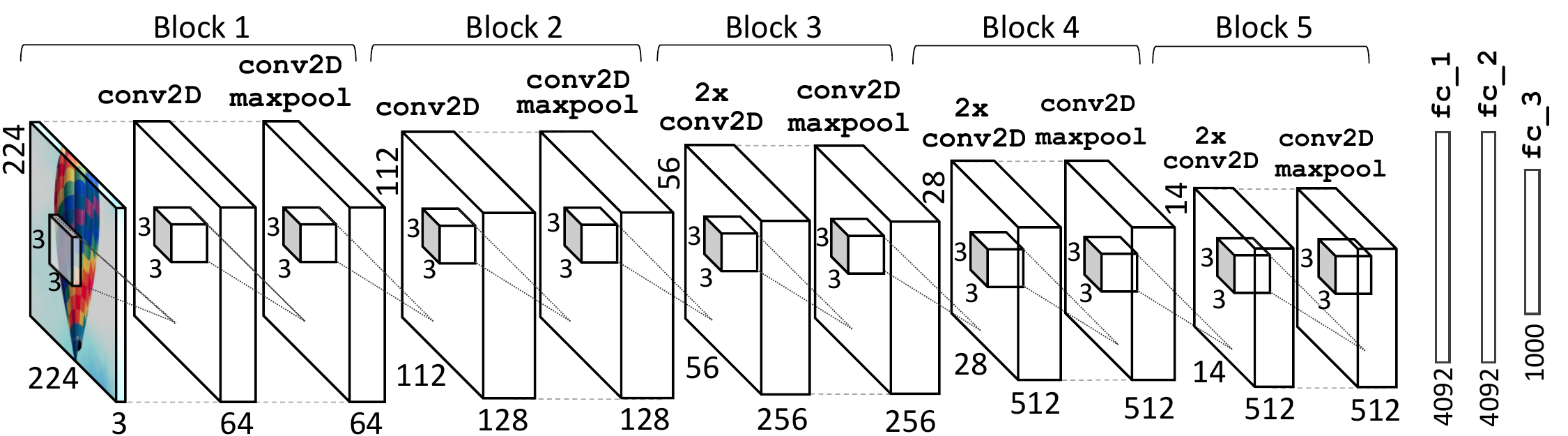}
\captionsetup{singlelinecheck=on,aboveskip=2pt, belowskip=0pt}
\caption{VGG16 model.}
\label{fig:vgg16}
\vspace{0pt}
\end{subfigure}
\vspace{5pt}

\end{tabular}

\vspace{-10pt}
\caption{Image recognition models.}  
\label{fig:image-models}
\vspace{-14pt}
\end{figure}

\noindent
Recent advancements in the deep visual recognition models led by the ImageNet large-scale visual recognition challenge (ILSVRC)~\cite{rus:den15} has allowed us to achieve high accuracies in various fields in computer vision, such as classification, detection, and segmentation. These models extensively use convolution neural network (CNN) layers and recent evidence reveals that the depth of a model is a critical factor in the accuracy~\cite{sze:liu15, sim:zis14-deep}. AlexNet~\cite{kri:sut12}, ZFNet~\cite{zei:fer14}, GoogLeNet\-~\cite{sze:liu15}, VGGNet~\cite{sim:zis14-deep}, and ResNet~\cite{he:zha15} are an example of these well-known models. Although such models have high accuracy and some of them surpass human-level accuracy~\cite{he:zha15}, their heavy computations are not ideal for resource-constrained devices~\cite{can:pas16}. Furthermore, this problem is worse when in real-time data processing. For demonstration, using Musical Chair, we studied AlexNet and VGG16, the model of which is shown in Figure~\ref{fig:image-models}.

\subsection{Action Recognition}
\label{sec:action-recognition}

\noindent
Recognizing human activities and classifying them (i.e., action recognition) in videos is a challenging task for DNN models. Since the type of actions is heavily correlated with motions, in comparison to models for still image classification, action recognition models, while performing still image classification, must also consider the temporal content in videos. In addition, action recognition models must tackle many other challenges, such as far-field recognition, limited resolution and low frame rate of videos. There has been a significant amount of attention in research community to build such models~\cite{sim:zis14, ryoo:kim17, jhu:ser07, kar:tod14, lap:mar08, wan:sch13}.

\subsubsection{Two-Stream CNN}
\label{sec:two-stream-head}

\noindent To date, the most accurate DNN models for action recognition~\cite{sim:zis14, ryo:rot17} utilize a particular baseline model based on two separate recognition streams with CNN layers, spatial and temporal. The outputs of which are combined in a temporal pyramid~\cite{cho:jeo08} and then fused in fully connected layers to produce outputs. The spatial stream classifies raw still frames for the video (i.e., images). On the other hand, the temporal stream processes a series of frames in a particular representation called optical flow~\cite{sim:zis14} and performs action classification from motions between frames. In the following subsections, using the paper of Ryoo et al.~\cite{ryoo:kim17} as a reference, we briefly describe each stream, optical flow representation, temporal pyramid, and implemented two-stream CNN model.

\begin{figure}[h]
\centering
\vspace{-5pt}
\includegraphics[width=1.0\linewidth]{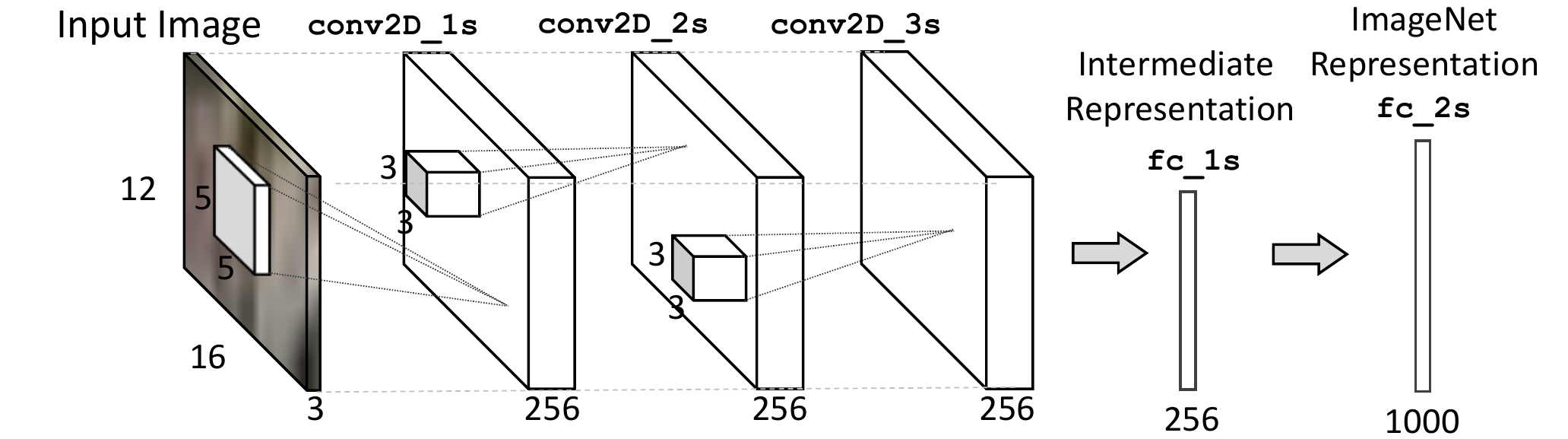}
\captionsetup{singlelinecheck=on,aboveskip=2pt, belowskip=0pt}
\caption{Spatial stream CNN.}
\label{fig:spatial}
\vspace{-7pt}
\end{figure}

\noindent \textbf{Spatial Stream CNN:}
\label{sec:spatial}
The spatial stream, similar to the competitors of the ImageNet challenge~\cite{rus:den15}, is implemented using convolution networks. An advantage of a separate stream for processing still images is the availability of huge datasets for training. Figure~\ref{fig:spatial} illustrates the spatial stream model in our implementation. The model, as input, takes a still image (i.e., a frame in a video) of size $16\x 12\x 3$ (in RGB), and processes this image with three convolution layers each with 256 filters, the kernel size of which are $5\x 5\x 3$, $3\x 3\x 256$, and $3\x 3\x 256$, respectively. Note that since we train this model based on the ImageNet dataset~\cite{rus:den15}, which has 1,000 different classes, the layer \texttt{fc\_2s} outputs 1,000 elements. However, in the action recognition model, we use the intermediate representation of the \texttt{fc\_1s} layer with a size of 256. Therefore, we reduce the dimensionality of the output, similar to the concept of word embedding~\cite{lev:gol14}, which is commonly used in neural text processing.

\begin{figure}[h]
\centering
\vspace{-5pt}
\includegraphics[width=1.0\linewidth]{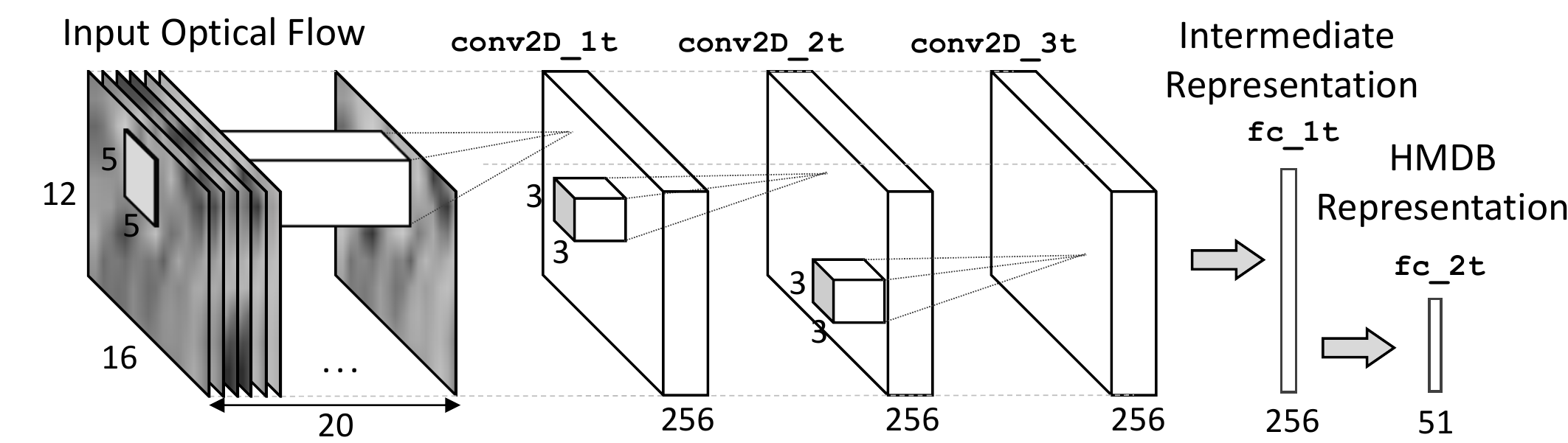}
\captionsetup{singlelinecheck=on,aboveskip=2pt, belowskip=0pt}
\caption{Temporal stream CNN}
\label{fig:temporal}
\vspace{-10pt}
\end{figure}

\noindent \textbf{Temporal Stream CNN:}
\label{sec:temporal}
The temporal stream CNN, shown in Figure~\ref{fig:temporal}, takes optical flow as input, which explicitly describes the motion between video frames. We use the F{\"a}renback~\cite{far:gun03} algorithm to find each pixel movement between two consecutive frames. In other words, for every pixel at a position $(u_t,v_t)$ at time $t$, the algorithm finds a displacement vector $\textbf{d}_t$ for each pair of consecutive frames, or $\textbf{d}_t=(d^x_t,d^y_t)=(u_{t+\Delta t}-u_t, v_{t+\Delta t}-v_t)$. In our temporal stream CNN, for 10 consecutive frames, we compute the optical flow and stack their $(d^x_t,d^y_t)$ to create an input with the size of $16\x 12\x 20$. Subsequently, the data is processed with three convolution layers each with 256 filters, the kernel size of which are $5\x 5\x 20$, $3\x 3\x 256$, and $3\x 3\x 256$, respectively. We train this temporal CNN with the HMDB dataset~\cite{kue:jhu11}. Therefore, the last fully connected layer (\texttt{fc\_2t}) in the temporal CNN has a size of 51. However, similar to the spatial CNN, in the action recognition model, the intermediate representation of the \texttt{fc\_1t} layer has a size of 256.

\begin{figure}[h]
\centering
\vspace{-10pt}
\includegraphics[width=1.0\linewidth]{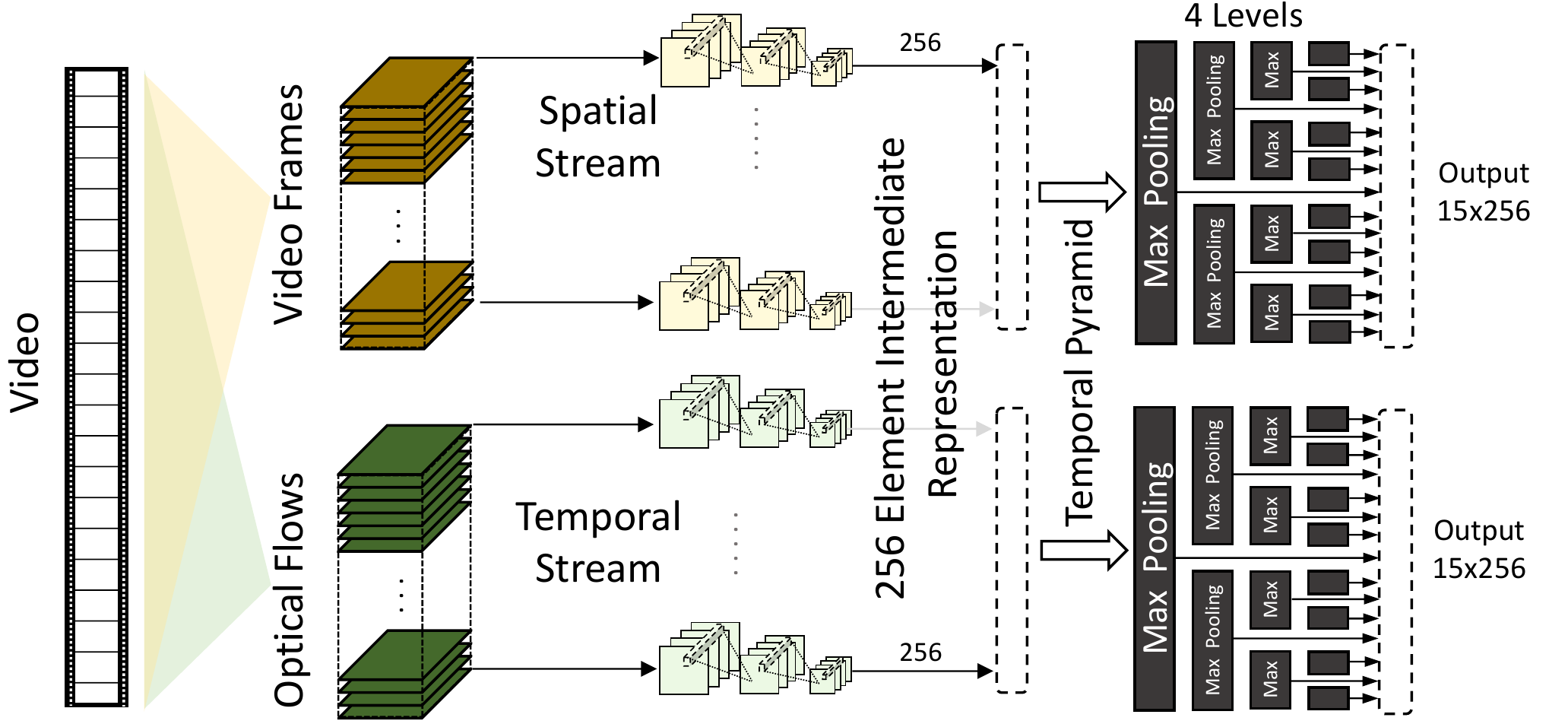}
\captionsetup{singlelinecheck=on,aboveskip=5pt, belowskip=0pt}
\caption{Four-level temporal pyramid generation in two-stream CNN.}
\label{fig:pyramid}
\vspace{-10pt}
\end{figure}

\noindent \textbf{Temporal Pyramid:}
\label{sec:pyramid}
To generate a single representation from the output of temporal and spatial streams, we generate a single spatio-temporal pyramid~\cite{cho:jeo08} representation for each video. The temporal pyramid creates an output with a fixed size that is agnostic to the duration of videos. Figure~\ref{fig:pyramid} depicts the steps of generating a four-level temporal pyramid from a video. First, in the spatial stream, each frame of the video is processed using the spatial CNN creating a group of 256-elements of intermediate representations. For the temporal stream, optical flows are calculated for each pair of consecutive frames, the intermediate representation of which is generated using the temporal CNN. Then, in each stream, 15 max-pooling layers with different input ranges generate a $15\x 256$ output. In other words, in each stream, the first max pooling layer applies max functions to all of the intermediate representations, the second and third max-pooling layers apply max functions to the first and second halves of the representations, respectively, and so on. By concatenating two $15\x 256$ outputs, we create the input of the final dense layer with a size $2\x 15\x 256$. 

\noindent \textbf{Final Dense Layers:}
\label{sec:two-stream-fc}
Finally, three dense layers with a size of 8192 (8k), 8192, and 51 perform the classification. Although the number of the layers in the final dense layer is smaller than that of the spatial or temporal CNNs, as we will see in Section~\ref{sec:nodes}, the computation and loading overhead of these layer are significant.

\noindent \textbf{Training Using Multi-Siamese CNN:}
A Siamese~\cite{siamese} neural network is a class of neural network architectures that contain two or more identical subnetworks. These subnetworks share the same parameters and weights, and parameter updating is mirrored in all subnetworks. The objective of such an architecture is to learn an embedding space that places similar items nearby~\cite{ryo:rot17, had:cho06, bel:bal15}. Our action recognition model utilizes a multi-Siamese CNN to effectively learn an embedding space to place similar videos nearby, while putting different videos far away. More specifically, we generate transformations of videos (i.e., scaling and rotation) and create positive and negative pairs for training. We use low-resolution transformations of the ImageNet~\cite{rus:den15} dataset, which contains 14 millions images in 1000 classes, to train the spatial CNN. Moreover, the HMDB dataset\-~\cite{kue:jhu11}, which has 7,000 videos in 51 classes, is used to train the temporal CNN. Then, we generate temporal pyramids and train the final fully connected layer. 

\vspace{-5pt}

\section{Distributing DNN}
\label{sec:parallel}

\noindent 
This section describes the methods of how Musical Chair distributes the computation of a model over multiple devices, and when it can parallelize computations between devices. Note that we are examining this problem in the context of real-time data processing, in which we have a continuous stream of data. For now, we assume each data input is independent (e.g., still image recognition). The ultimate goal is to reduce the effective process time per input data. In this paper, we call the processes that are performed on an input data \emph{tasks}. A \emph{task} is the process performed by a layer or a group of consecutive layers. We are particularly interested in understanding the differences between \emph{data parallelism} and \emph{model parallelism} that is applicable to a \emph{task}. (The terms data and model parallelism are inspired by concepts in GPU training of DNNs~\cite{coa:huv13}.) We define data parallelism as duplicating devices that perform the same task, or share the same model parameters. By doing so, we increase the throughput of the system. For instance, by adding another device that performs the same task on a different input, in an optimal case, the throughput doubles. On the other hand, in model parallelism, we distribute a task, which is dividing the task to subparts and assigning different subparts of the \emph{same} task to additional devices. The computation of subparts, depending on the arrangement of layers in a task, is either parallel or sequential. In model parallelism, since the parameters of the model are divided between devices, the parameters are not shared. In summary, in data parallelism, devices share the same model, whereas in model parallelism, devices share the same data input. 

\begin{figure}[t]
\centering
\vspace{-8pt}
\includegraphics[width=1.0\linewidth]{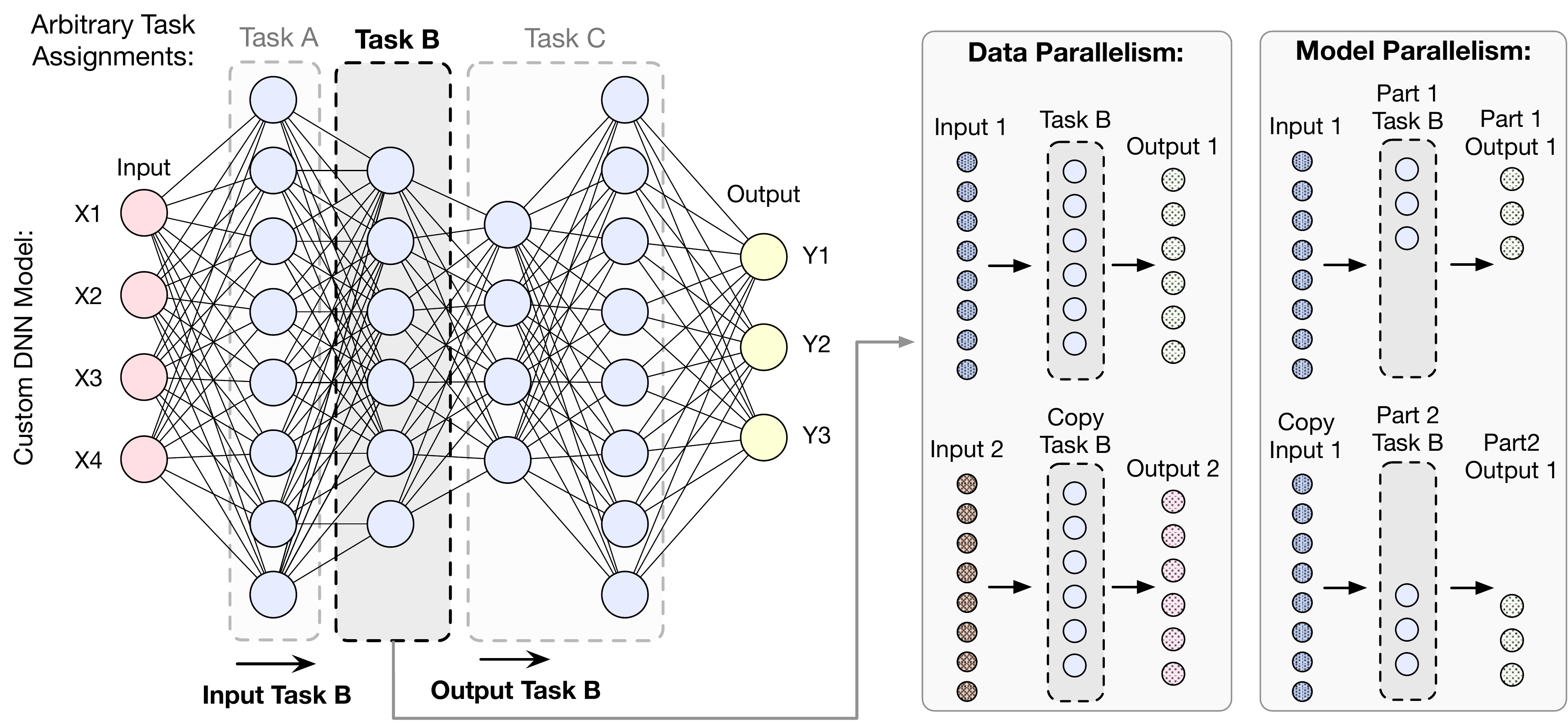}
\captionsetup{singlelinecheck=on,aboveskip=5pt, belowskip=0pt}
\caption{Model and data parallelism for task B on two devices. Real-time data provides us independent inputs.}
\label{fig:model-data-concept}
\vspace{-15pt}
\end{figure}

Figure~\ref{fig:model-data-concept} depicts model and data parallelism of task B, an arbitrary task, in an example DNN with five layers. The illustration shows that, for two devices, data parallelism basically performs same Task on two independent inputs. In model parallelism, one input is fed to two devices that perform half of the computation. To create the final output, or the input of the next task, a merge operation is required. For simplicity, we assume inputs are independent, which is not always true. For instance, in the action recognition model described in Section~\ref{sec:two-stream-head}, the temporal CNN and temporal pyramid input is a concatenation of several frames or computations of previous layers. We address this challenge with a sliding window concept in Section~\ref{sec:sliding-window}. In real-time applications, implementing data parallelism is basically assigning each newly arrived data, for instance in a round-robin fashion, to the devices. However, performing model parallelism requires a knowledge of deep learning. In fact, the effectiveness of model parallelism depends on various factors such as the type of layers, input and output size, and amount of data transfer. In addition, the performance is tightly coupled with the balance of the computation across devices, whereas, in data parallelism, since each device performs the same computations, the computations are inherently balanced. We study the application of model parallelism for \texttt{fc} and \texttt{conv} layers since the remaining layers, mentioned in Section~\ref{sec:ML-layers}, perform element-wise operations (easily parallelizable) or are not computationally intensive.

\subsection{Fully Connected Layer}

\noindent 
In a fully connected layer, the value of each output is dependent on the weighted sum of all inputs. To apply model parallelism to an \texttt{fc} layer, we can distribute the computation of each output while we transmit all input data to all devices. Since the model remains the same, such distribution does not require training new weights. However, all the input values are copied to each device. Later, when each subcomputation is done, we need to merge the results for the input of the next layer. Another approach is to provide a subset of the input data to each output. In this approach, the communication overhead of copying a subset of input data is less than that of copying all of the input. This approach creates new DNN models, and for each model retraining is necessary to learn a new set of weights. Hence, since copying all the input data uses the same model while reducing the amount of computation per device, Musical Chair utilizes this approach.

\begin{figure}[b]
\centering
\vspace{-15pt}
\includegraphics[width=1.0\linewidth]{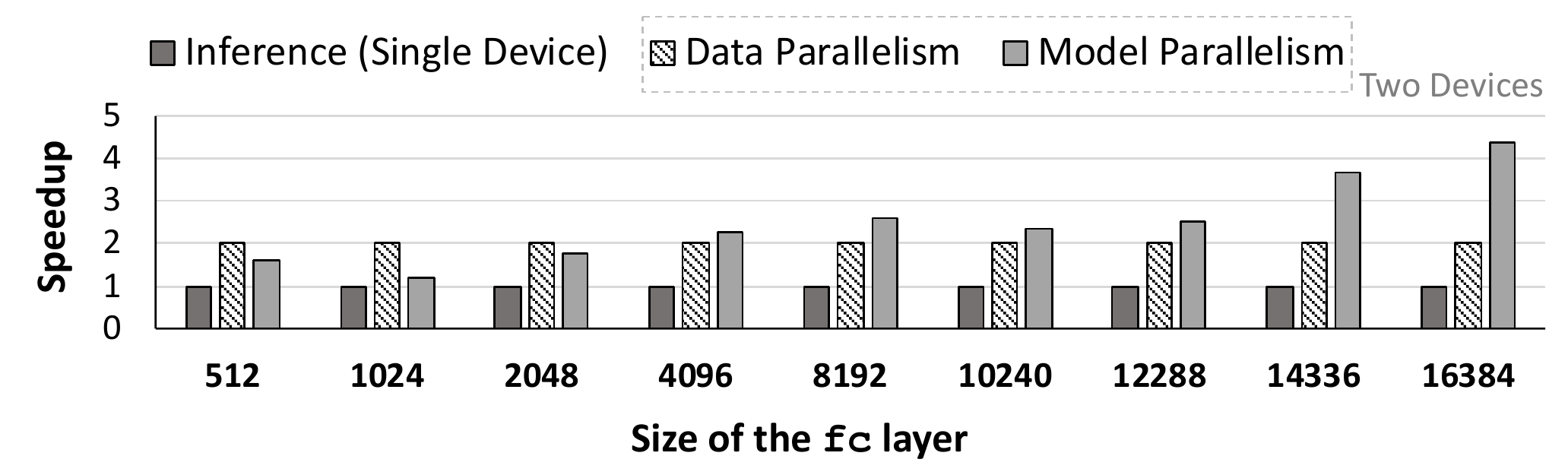}
\captionsetup{singlelinecheck=on,aboveskip=2pt, belowskip=0pt}
\caption{Performance speedup of model and data parallelism on two Raspberry PIs executing an \texttt{fc} layer.}
\label{fig:model-vs-data}
\vspace{-5pt}
\end{figure}

\begin{figure}[!b]
\centering
\vspace{-5pt}
\includegraphics[width=1.0\linewidth]{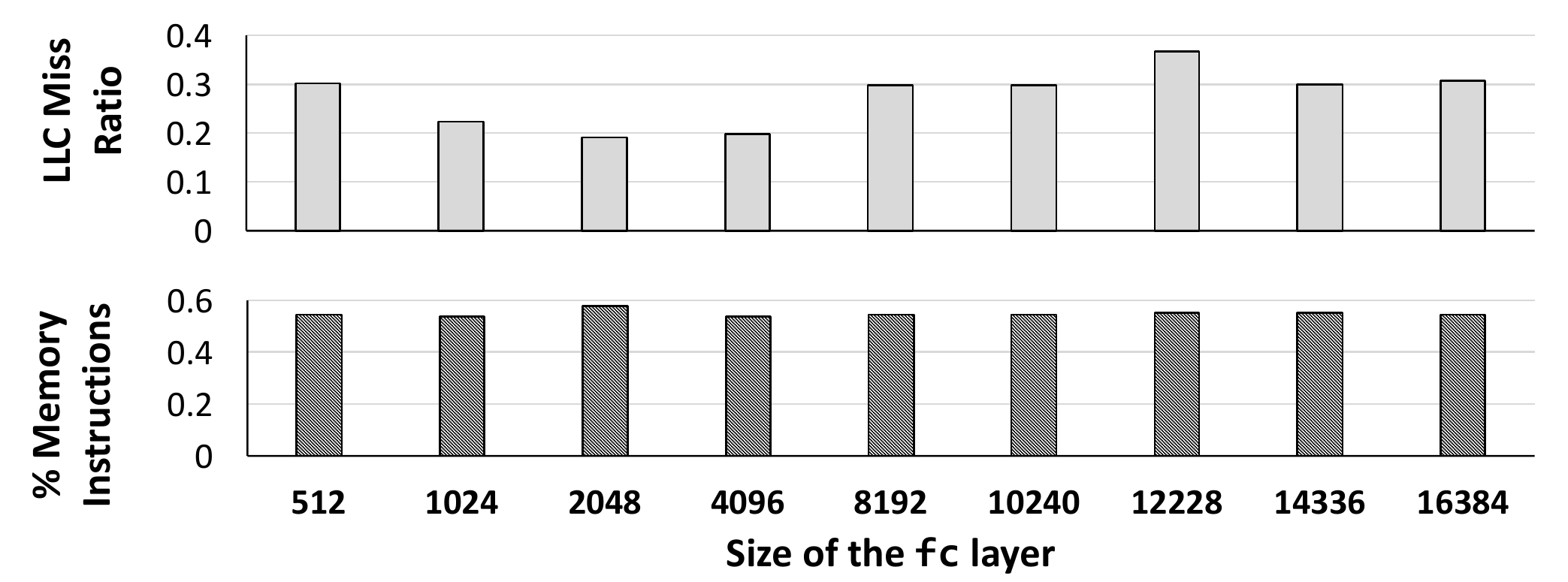}
\captionsetup{singlelinecheck=on,aboveskip=2pt, belowskip=0pt}
\caption{Gathered performance counters during the \texttt{fc} layer execution.}
\label{fig:model-vs-data-details}
\vspace{-5pt}
\end{figure}

As an example of how model and data parallelism affect the performance, we examine various \texttt{fc} layers, the input size of which are 7,680, but with different sizes. For each layer, we measure its performance (i.e., throughput) on a Raspberry PI, the specification of which is outlined in Table~\ref{tab:pi}. Figure~\ref{fig:model-vs-data} illustrates performance speedup (i.e. throughput) of model and data parallelism for each size normalized to performing inference on a single device. Because we add a similar device to the system that performs exactly the same task, data parallelism has almost twice the performance of the baseline. On the other hand, the performance of model parallelism is dependent on the performance of a half-sized \texttt{fc} layer on the PI. This is because in model parallelism we divide an \texttt{fc} layer of size $y$ into two \texttt{fc} layers of size $\nicefrac{y}{2}$. As Figure~\ref{fig:model-vs-data} shows, in some \texttt{fc} layers, model parallelism performs better than data parallelism. Note that since both models have the same input size (Figure~\ref{fig:model-data-concept}), the difference in communication overhead, which is due to difference in output size and merging the data, has minor noticeable effects in both models. As we will see, for complex systems, with several nodes, this overhead will reduce the potential performance gain. 
 To gain insights, we collect ARM Cortex-a53~\cite{a53} performance counters, depicted in Figure~\ref{fig:model-vs-data-details}. We observe all \texttt{fc} layers have a high percentage of memory instructions, 50\%, and their last level cache (LLC) miss ratio is significantly high. In fact, for \texttt{fc} layers larger than 10,240, processor starts using the memory swap space. Since in model parallelism a layer is distributed on more than one device, we avoid swap space activities which results in speedups greater than 2$\x$. On a resource-constrained device, these facts combined with limited parallelism of the \texttt{fc} layer and a high amount of computation, allows a higher gain in model than data parallelism. In fact, next section will describe how \texttt{conv} layers always have better performance with data parallelism.

\subsection{Convolution Layer}
\noindent 
In a convolution layer, several filters are swept over the input data. In order to parallelize a \texttt{conv} layer, we can (\romannum{1}) distribute filters while copying the input, (\romannum{2}) divide input while copying all filters, or (\romannum{3}) a mix of the last two options. The efficiency of each of these solutions depends on the number of filters, size of each filter, and dimensions of the input. Generally, similar to our model, the dimensions of the input is not significantly large, but the number of filters is large. Hence, distributing filters among devices is common. Furthermore, we need to merge all results to create the input to the next layer, so this method of parallelization is only applicable to a single layer of CNNs. 

\begin{figure}[t]
\centering
\vspace{-10pt}
\includegraphics[width=1.0\linewidth]{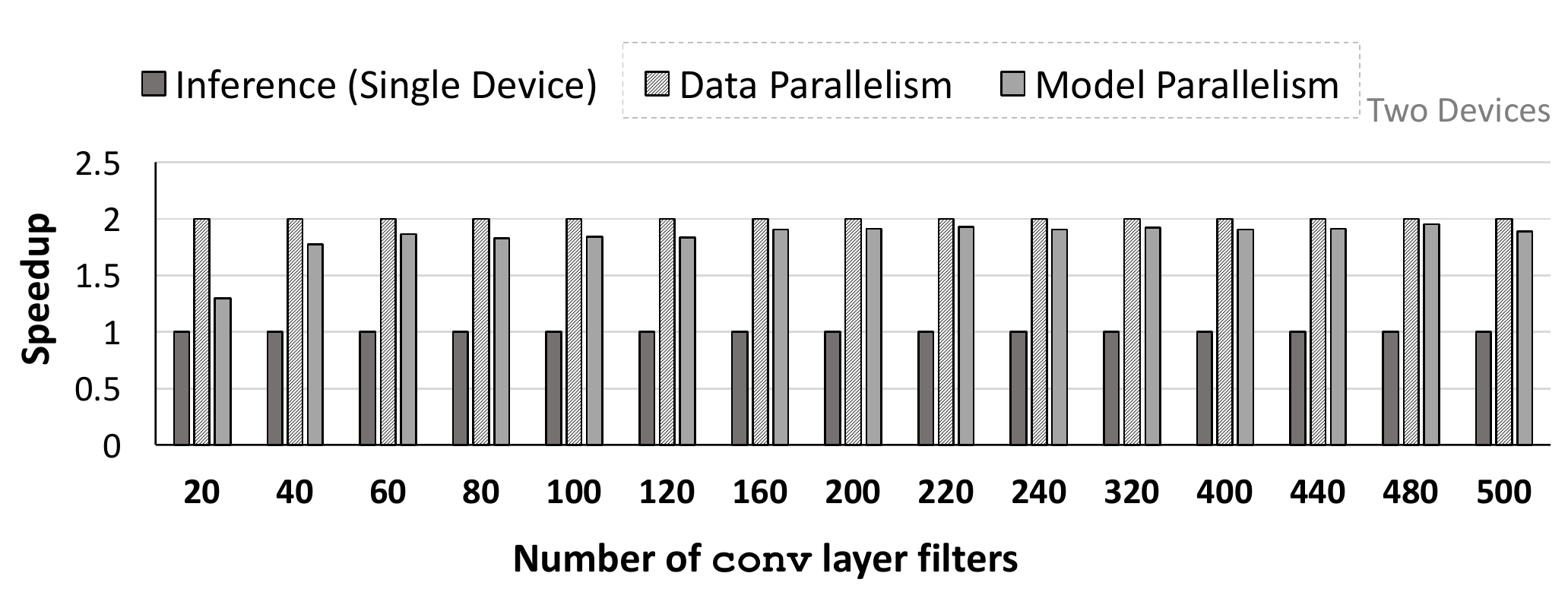}
\captionsetup{singlelinecheck=on,aboveskip=2pt, belowskip=0pt}
\caption{Performance speedup of model and data parallelism on two Raspberry PIs executing a \texttt{conv} layer.}
\label{fig:model-vs-data-cnn}
\vspace{-5pt}
\end{figure}

\begin{figure}[!t]
\centering
\vspace{-10pt}
\includegraphics[width=1.0\linewidth]{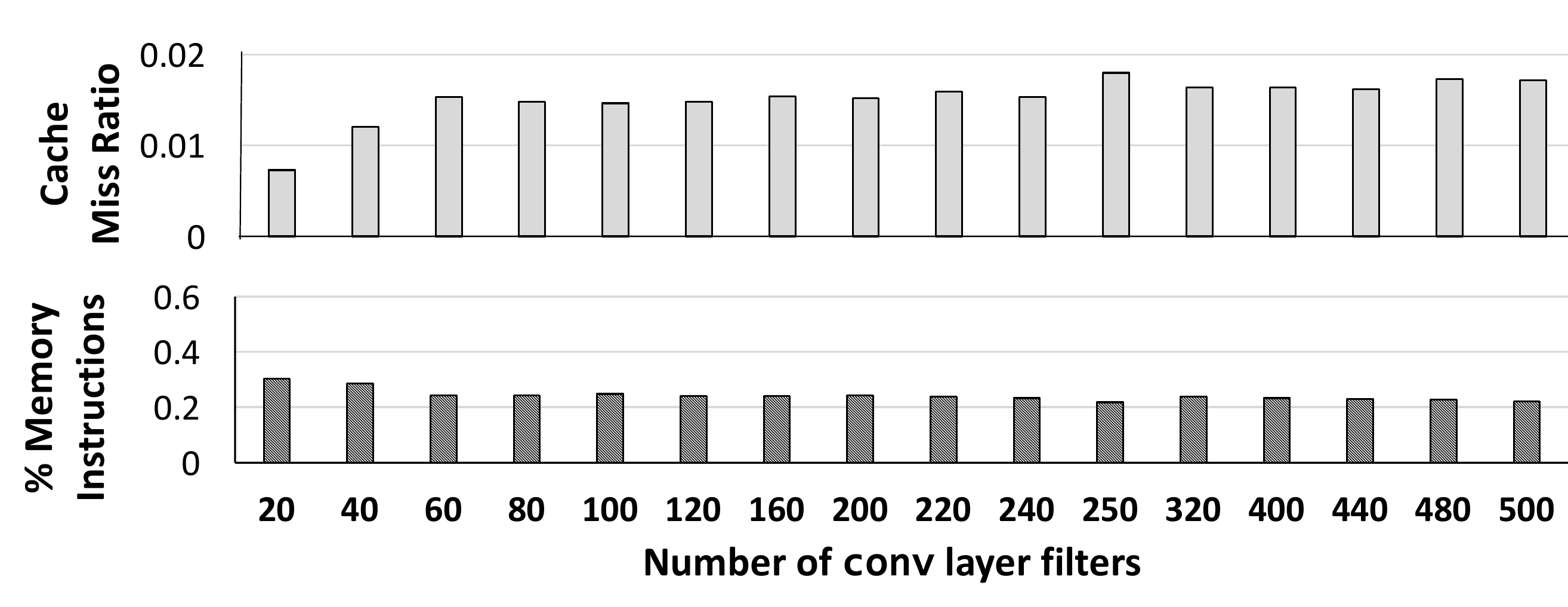}
\captionsetup{singlelinecheck=on,aboveskip=2pt, belowskip=0pt}
\caption{Gathered performance counters during the \texttt{conv} layer execution.}
\label{fig:model-vs-data-details-cnn}
\vspace{-17pt}
\end{figure}

As an example, we examine a series of CNN layers with the input size of 200$\x$200$\x$3 ($\simeq$ 0.5\,MB) and the kernel size of 5$\x$5, with a different number of filters (i.e., kernels). Similar to the \texttt{fc} layer study, we measure the performance of model and data parallelism of these layers on a Raspberry PI. Figure~\ref{fig:model-vs-data-cnn} illustrates that data parallelism always has a better performance than that of model parallelism. One reason is that in model parallelism we have to transmit the \emph{same} input, while in data parallelism we transmit different inputs. In other words, model parallelism wastes the half of the data communication bandwidth by sending the same data into two devices. Furthermore, from the performance counters in Figure~\ref{fig:model-vs-data-details-cnn} we see that, in comparison with \texttt{fc} layers, \texttt{conv} layers have a lower cache miss ratio\footnote{since the \texttt{conv} layer mostly fits into the first- level cache, we show its miss ratio instead of LLC miss ratio.} and less percentage of memory instructions.\texttt{conv} layer computations have more locality because the size of each filter is small, and all filters share the same input. On the other hand, similar reasoning applies when the size of each filter is large because we can efficiently divide the input to create smaller workload size. In addition, the \texttt{conv} layer has more parallelism, because the computation of each filter is independent. Based on these gained insights, we propose our solution for distributing DNNs in the next section.  

\begin{figure*}[h]
\centering
\vspace{0pt}
\includegraphics[width=1.0\linewidth]{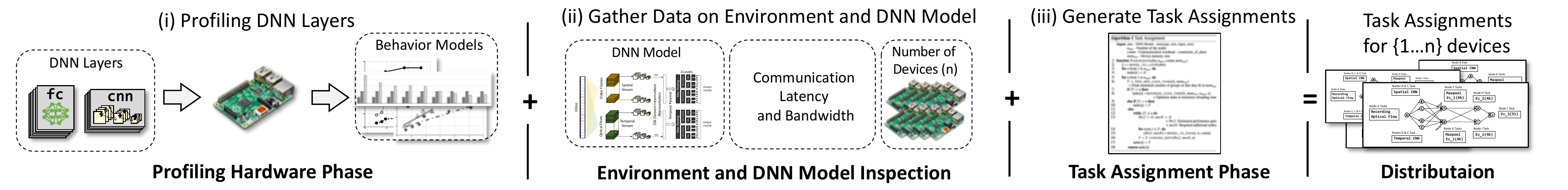}
\captionsetup{singlelinecheck=on,aboveskip=5pt, belowskip=0pt}
\caption{Steps for generating task assignments in Musical Chair.}
\label{fig:solution}
\vspace{-15pt}
\end{figure*}

\subsection{Musical Chair Solution}
\label{sec:music-chair-solution}

\noindent 
To find the optimal distribution for each DNN model, given the number of devices in the system, we devise a solution based on profiling. As explained in the previous section, profiling is necessary for understanding the performance benefits of data and model parallelism. In the Musical Chair solution, shown in Figure~\ref{fig:solution}, first we profile different DNN layers, similar to the previous section. Furthermore, we also measure the memory usage (more details in Section~\ref{sec:simple-systems}). The profiling is performed once for creating behavior models of DNN layers. Therefore, after creating the behavior models, no additional cost is associated with profiling. In the second step, our solution considers the target DNN model, number of devices, and communication overhead. In this paper, we thoroughly analyze a state-of-the-art action recognition model, and two image recognition models as well.

In a distributed system, achieving high performance and real-time inference requires careful load balancing between the devices. First, we define performance in a real-time inference flow. From a user perspective, performance is summarized in two factors: (\romannum{1}) the number of performed inferences per second, or $IPS$, and (\romannum{2}) the delay of the system in recognizing a new input, or $t_{forward}$. Musical Chair aims at increasing $IPS$ in a distributed system while keeping $t_{forward}$ within a specified range. $t_{forward}$ is influenced by three system-level criteria: (\romannum{1}) If any resource-constrained device in the system performs more than one task, it needs to load different models during real-time processing. Such loading time increases $t_{forward}$ significantly as we will discuss in Figure~\ref{fig:loading-time}. (\romannum{2}) If any task in the system is performed in parallel (data or model parallelism). In such a scenario, the inference time of that task is reduced, dependent on the number of parallel devices, and the overhead of communication and data management. (\romannum{3}) A system with limited computational power cannot have $t_{forward}$ less than its theoretical optimum, dependent on the computational power of its devices. In addition, the maximum $IPS$ of a system is limited by the execution time of the slowest task. To increase $IPS$, we have a few options: (\romannum{1}) increase the computational power of the slowest device, (\romannum{2}) split the task assigned to the device using model parallelism, and (\romannum{3}) parallelize the task among some devices using data parallelism. All of the mentioned techniques perform a version of load balancing in our distributed system to increase $IPS$.

\vspace{-5pt}
\begin{algorithm}[h]
   \small
   \caption{Task Assignment Algorithm.}
   \label{algo:algo}
    \begin{algorithmic}[1]
    
      \begin{flushleft}
        \textbf{Input:} $dnn$ - DNN Model - $dnn$(type, size, input\_size) \\
        \hspace{22pt} $n_{max}$ - Number of the devices \\
        \hspace{22pt} $comm$ - Communication overhead - $comm$(size\_of\_data) \\
        \hspace{22pt} $mem_{size}$ - Device memory size
      \end{flushleft}
      \Function{TaskAssign}{$dnn, n_{max}, comm, mem_{size}$}         
          \State $L$ $\coloneqq$ \Call{model\_to\_layer}{$dnn$} 
          \item[] \Comment{Also consider data dependency between layers} 
          \For{$n$ from $1$ to $n_{max}$:}
            \State tasks$[n]$ $\coloneqq$ $\varnothing$
          \EndFor
          \For{$n$ from $1$ to $n_{max}$:} \label{line:main_for}
            \State $T$ $\coloneqq$ \Call{find\_min\_load\_tasks}{$L$, $mem_{size}$} \label{line:aglo:find_min_load_tasks}
            \item[] \Comment{Return the minimum number of groups as a set that fit in $mem_{size}$}
            
            \If{$|T| > n$}
              \State tasks$[n] = $\Call{minimize\_load\_time}{$T$, $mem_{size}$, $n_{max}$, $n$} \label{line:aglo:greater}
              \item[] \Comment{Optimize and regroups tasks to minimize reloading time}
            \ElsIf{$|T| = n$}
              \State tasks$[n] = T$ \label{line:aglo:equal}
            \Else
             \While{$|T| \neq n$}
             \State $Perf$ $\coloneqq$ $\varnothing$, $newN$ $\coloneqq$ $\varnothing$
             \item[] \Comment{$Perf$: Estimated perfomance gain}
             \item[] \Comment{$newN$: Required additional devices}
             \For{every $t \in T$:}
                \State [$Perf$, $newN$] = \Call{model\_vs\_data}{$t$, $n$, $comm$} \label{line:aglo:less}
             \EndFor
             \State [$t_{old}, t_{new}]$ = \Call{choose\_best}{$Perf$, $newN$, $n$} \label{line:aglo:best}
             \State $T = T - t_{old} + t_{new}$ 
             \EndWhile
             \State tasks$[n] = T$
            \EndIf
            \State \Return tasks$[n]$
          \EndFor
      \EndFunction
      
\end{algorithmic}
\vspace{0pt}
\end{algorithm}
\vspace{-5pt}

In step three (Figure~\ref{fig:solution}), using profiled data, we generate task assignments based on the flow of Algorithm~\ref{algo:algo}. Because of the possibility that during execution some devices are inactive, busy, or has more than one input, Musical Chair generates task assignments for one device to the total number of devices in the system (Line~\ref{line:main_for}). Since we have all of the task assignments for any number of devices, our system can dynamically change the number of collaborating devices. The algorithm also accounts for the overhead of communication when making decision using the input profiled data in $comm$ variable. In line~\ref{line:aglo:find_min_load_tasks}, with a given memory size per device ($mem_{size}$), the algorithm finds the minimum number of tasks so that each task fits in memory. Then, based on the number of tasks($|T|$) and the number of devices ($n$), the algorithm decides to reduce, increase, or return the tasks. If $|T|$ equals $n$, the algorithm returns the tasks computed in line~\ref{line:aglo:find_min_load_tasks}. If $|T|$ is greater than $n$, the algorithm, in line~\ref{line:aglo:greater}, regroups some tasks while optimizing the memory reloading overhead. Else, If $|T|$ is less than $n$ (line~\ref{line:aglo:best}), the algorithm uses the gained insights from data and model parallelism to create the most optimized distribution for $n$ devices.

\section{Musical Chair for \\Action Recognition}
\label{sec:music-act}

\noindent 
This section thoroughly examines real-time DNN-based action recognition, its requirements, and gained insights for resource-constrained IoT devices for Musical Chair.

\subsection{Simple IoT Devices}
\label{sec:nodes}

\noindent 
Musical Chair is designed to harvest the computation power of simple and widespread IoT devices. In this paper, as a case study, we utilize several Raspberry PIs~\cite{pi3}, the specification of which is in Table~\ref{tab:pi}. The Raspberry PI is a cheap, low power, and affordable device that is a realistic or even less computationally powerful example of today's some IoT devices such as the one in the Nest Thermostat~\cite{nest-term-breakdown,nest-term-breakdown-arm,nest-6core}. On each PI, with the Ubuntu 16.04 operating system, we use Keras 2.0~\cite{chollet2015keras} with the TensorFlow 1.0~\cite{tensorflow2015-whitepaper} backend. To measure the power consumption of a single PI, we use a USB digital multimeter. To measure the power consumption of a system with PIs, we power all PIs with a 14-port powered USB 3.0 hub, while measuring its power consumption. Reported idle power and 100\% utilization power consumption of one PI is 1.3\,W and 6.5\,W, respectively. Please note that since we do not use all capabilities of a single PI, such as the display, GPIO, ADC, and DAC, the idle power is higher than just power of the processors.  The averaged observed power also includes communication and memory accesses.

\renewcommand{\arraystretch}{0.9}
\begin{table}[h]
    \small
	\centering
	\vspace{-5pt}
	\captionsetup{singlelinecheck=on,aboveskip=1pt}
	\caption{Raspberry PI 3 specification~\cite{pi3}}
	\begin{tabular}{c | c | c}
		\toprule
        CPU & \multicolumn{2}{c}{1.2\,GHz Quad Core ARM Cortex-A53} \\
        Memory &  \multicolumn{2}{c}{900\,MHz 1\,GB RAM LPDDR2} \\
        GPU & \multicolumn{2}{c}{No GPGPU Capability} \\
        Price & \multicolumn{2}{c}{\$35 (Board) + \$5 (SD Card)} \\
        \midrule
        \multirow{3}{2cm}{\centering Power Consumption }
        &Idle (No Power Gating) & 1.3\,W \\
        &\%100 Utilization & 6.5\,W \\
        & Averaged Observed & 3\,W \\
		\bottomrule
	\end{tabular}
	\label{tab:pi}
	\vspace{-5pt}
\end{table} \renewcommand{\arraystretch}{1}

\subsection{Simple Systems}
\label{sec:simple-systems}

\begin{figure}[t]
\centering
\vspace{-0pt}
\includegraphics[width=1.0\linewidth]{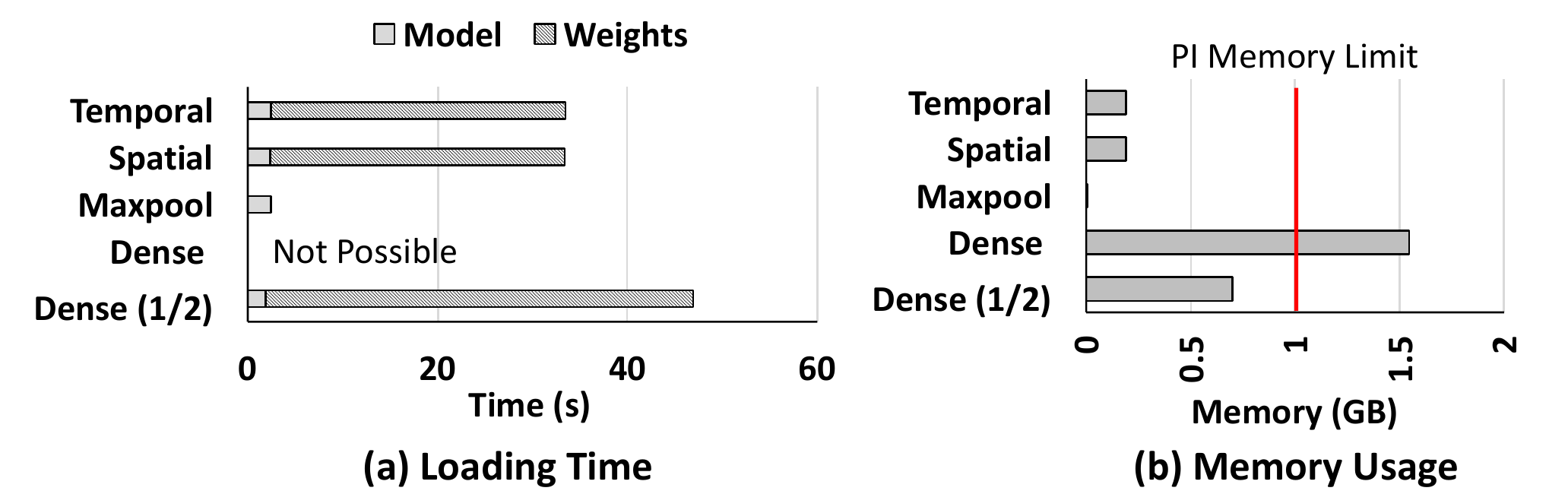}
\captionsetup{singlelinecheck=on,aboveskip=5pt, belowskip=0pt}
\caption{The loading time (a) and memory usage (b) of general tasks in action recognition on a Raspberry PI.}
\label{fig:loading-time}
\vspace{-15pt}
\end{figure}

\begin{figure}[h]
\centering
\vspace{-5pt}
\includegraphics[width=1.0\linewidth]{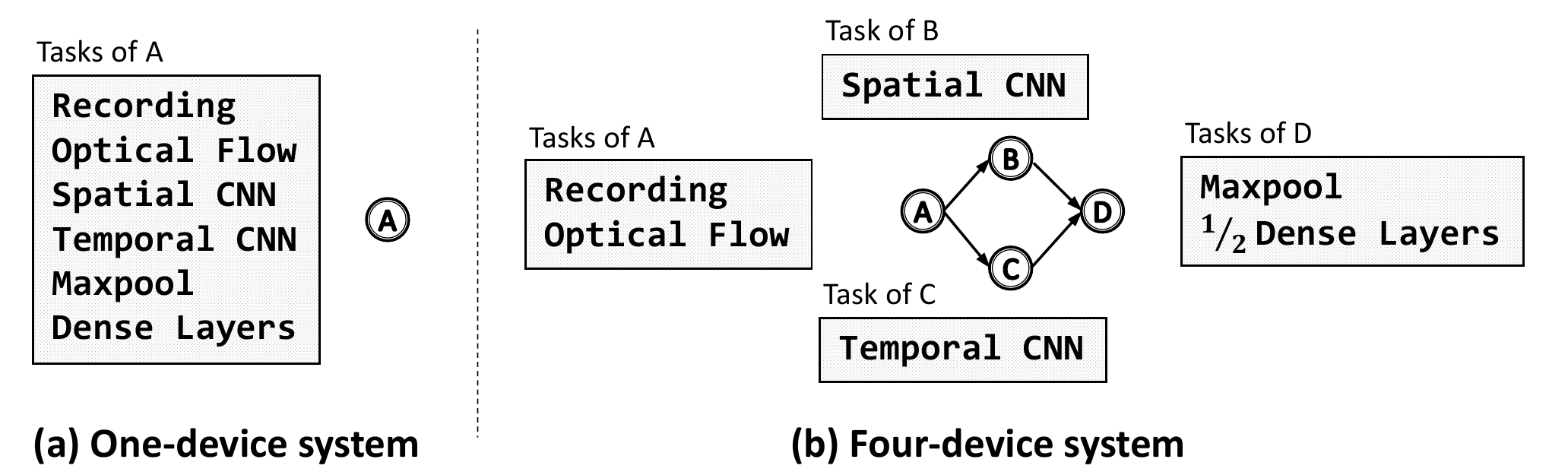}
\captionsetup{singlelinecheck=on,aboveskip=5pt, belowskip=0pt}
\caption{Simple systems with general tasks: (a) one-device, and (b) four-device system.}
\label{fig:nodes-simple}
\vspace{-5pt}
\end{figure}

\noindent 
Since Musical Chair dynamically adjusts the number of devices, and the task of each device, we have to load all trained weights of the model to the storage of each PI in the initialization. However, one PI cannot handle executing or loading all of the weights to its memory. In details, Figure~\ref{fig:loading-time} shows the loading time and memory usage of the general tasks (i.e., temporal CNN, spatial CNN, temporal pyramid, maxpooling, and dense layers) in the action recognition model. Since the memory usage of the original dense layer is larger than 1\,GB, a single PI cannot efficiently execute the whole dense layer. To understand the requirements of similar dense layers, we measure the time and memory usage of a dense layer with half-sized dimensions of the original one, which has a 15\% lower accuracy than the original dense layers. As Figure~\ref{fig:loading-time}, even if the computation time of each task was negligible, in one device system, shown in Figure~\ref{fig:nodes-simple}a, the overhead of loading each task would be notably high for meaningful real-time processing.

\begin{figure}[h]
\centering
\vspace{-5pt}
\includegraphics[width=1.0\linewidth]{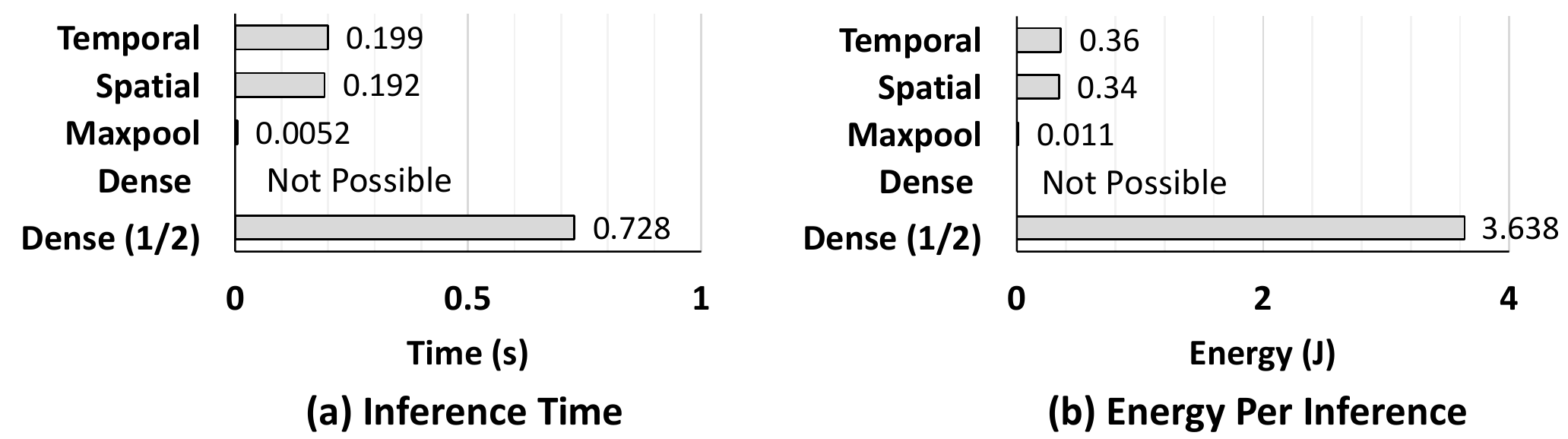}
\captionsetup{singlelinecheck=on,aboveskip=5pt, belowskip=0pt}
\caption{The time (a) and energy (b) per inference of general tasks in action recognition on a Raspberry PI.}
\label{fig:inference-time}
\vspace{-15pt}
\end{figure}

To overcome the overhead of loading each task repeatedly, consider a case with four devices, each of which handles a set of the general tasks. As shown in Figure~\ref{fig:nodes-simple}b, since each task is loaded once, real-time data processing is more efficient than a one-device system. However, as Figure~\ref{fig:inference-time} shows, the inference time of half-sized dense layer is more than 0.7 seconds, while its energy per inference is $10\x$ larger than that of spatial or temporal CNNs. Hence, in such an implementation, we still cannot efficiently process real-time data since even the computation of the half-size dense layer creates a bottleneck. Furthermore, to implement such a half-size dense layer, we must retrain the model. Note that the provided tasks are shown for illustrating the challenges of implementing DNNs on resource-constrained devices. In fact, Musical Chair, as we will see, solves the presented bottleneck of dense layers with an additional device.

\subsection{Communication}
\label{sec:comm}

\noindent
\textbf{Hardware:} We use a local Wifi network in our experiments, the measured bandwidth of which is 62.24\,Mbps, with a measured client-to-client latency of 8.83\,ms for 64\,B. Since Musical Chair needs a model for communication overhead between devices, we measured such overhead by fitting a line over measured data with the equation $t = 0.0002d+0.002$, in which $t$ is end-to-end latency (seconds) and $d$ is the size of data (kB). Moreover, our energy numbers \emph{includes} communication energy for devices.

\noindent
\textbf{Basic Communication:}
In Musical Chair, devices need to communicate with each other efficiently for transmitting data and commands. In this section, we overview the communication protocol deployed in our system. We integrate Apache Avro~\cite{apache}, a remote procedure call (RPC) and data serialization framework with Keras in Musical Chair. The RPC capacity of Avro enables Musical Chair to request a service from a program located in another device on the local network without providing the details of the network specification. In addition, Avro's data serialization capability provides flexible data structures for transmitting and saving data during processing while preserving its format (e.g., string or float). Therefore, a device may offload the results of a computation to another device (e.g., the results from spatial stream CNN on a frame) and initiate a new process. In our implementation, Avro messages are one-way data transfers and include fields for data (e.g., data, image, or optical flow), next device, and source device (i.e., current sending device, which is useful when we combine processed data together).

\begin{figure}[h]
\centering
\vspace{-5pt}
\includegraphics[width=1.0\linewidth]{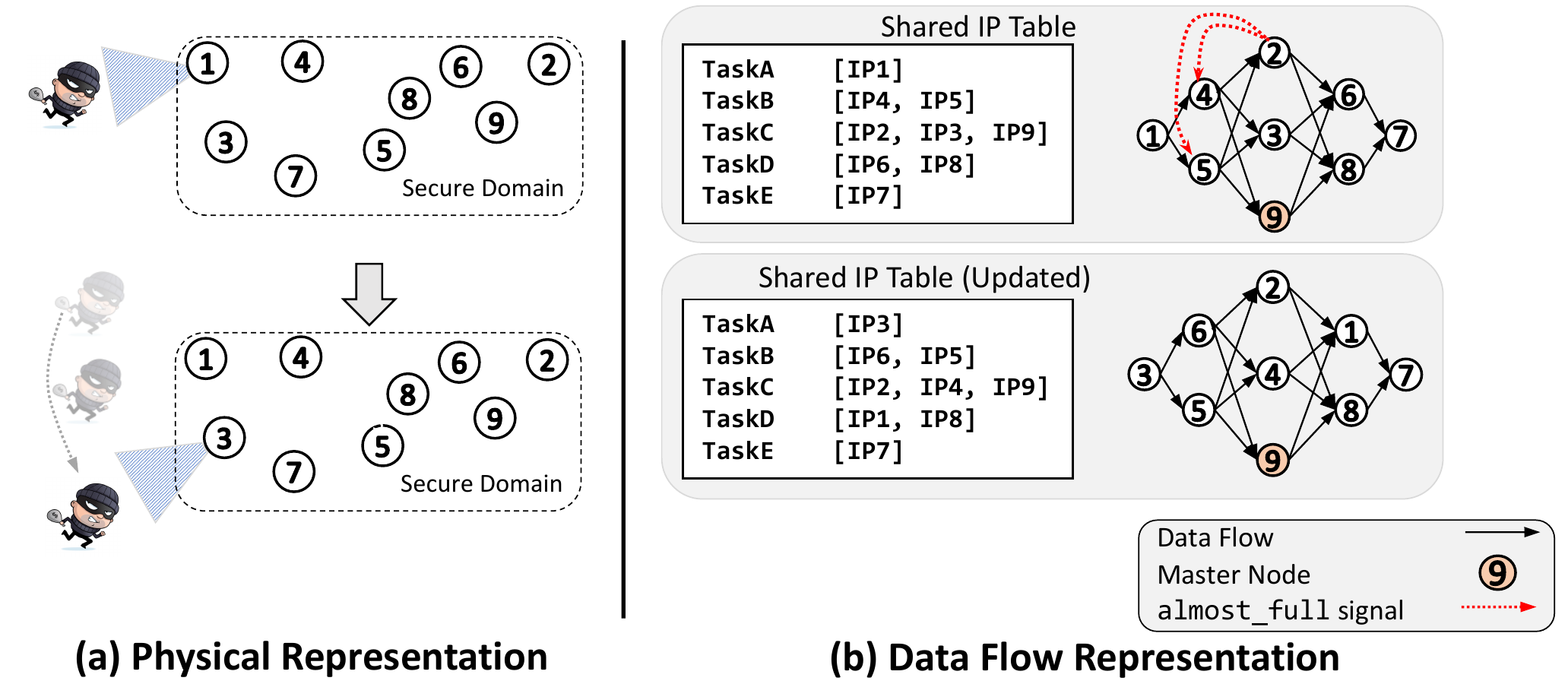}
\captionsetup{singlelinecheck=on,aboveskip=5pt, belowskip=0pt}
\caption{Example of dynamic communication flow and IP table updates.}
\label{fig:musical-chair}
\vspace{-20pt}
\end{figure}

\noindent
\textbf{Dynamic Communication Flow:}
To effectively identify all devices, each device has a local copy of a shared IP address table. In this IP table, each device is designated to perform a specific task. Therefore, with a knowledge of the general tasks of a model, each device identifies its previous and next set of devices, $G_p$ and $G_n$, respectively.  Moreover, if the condition of the environment is changed, for instance, the object has moved into the field of view of another device, a musical chairs game\footnotemark~will be triggered, and the task of devices in the IP table is updated. To avoid conflicts, an update to the IP table is only performed by a master device, which is chosen randomly among available devices during system initialization. Similarly, all communication regarding task assignments and condition updates of the environment is only between slave devices and the master. In order to create a dynamic data flow in the system, each device in Musical Chair maintains a fixed size buffer for incoming data. Whenever the buffer is almost full, the maintaining device sends an \texttt{almost\_full} signal to all its previous devices ($G_p$), which permits them to drop some input data (i.e., reducing sampling frequency)  (this is done without disturbing pending sliding window data, see Section~\ref{sec:sliding-window}). Figure~\ref{fig:musical-chair} shows an example network, its data flow representation, and how the shared IP table is updated. As shown, when the object exits the field of view of device one and enters into device three's, the master device updates the tasks and data flow. In addition, an example of sending the \texttt{almost\_full} signal from device two to its previous devices is shown. 

\footnotetext{``Musical chairs is a game of elimination; when the music stops whichever player fails to sit on a chair is eliminated'' (Wikipedia)}

\subsection{DNN Real-Time Processing}

\noindent
State-of-the-art DNN models are not designed or optimized for real-time processing on mobile devices. In other words, we either need a powerful server or have to rely on the cloud for a meaningful recognition experience in real time~\cite{kan:hau17, can:pas16}. Similarly, the model presented in Section~\ref{sec:two-stream-head} is designed for offline action recognition, in which videos are saved, and then processed offline. To perform recognition for this model in real-time, using Musical Chair, we distribute computations effectively among devices and utilize a sliding window over acquired data.

\ignore{
{\color{red}maybe update with Tegra measured case}
State-of-the-art DNNs are designed to maximize the accuracy of the models, which naturally leads to deeper and resource-hungry models. For instance, most of the accurate and recent DNNs for still image recognition are not suitable for real-time processing on the NVIDIA Jetson TX1 board~\cite{jetson}, which has a one TFLOPS 256-core Maxwell GPU~\cite{can:pas16}. These models are not designed or optimized for real-time processing on mobile devices. In other words, we either need a powerful server or have to rely on the cloud for a meaningful recognition experience in real time~\cite{kan:hau17, can:pas16}. Similarly, the model presented in Section~\ref{sec:two-stream-head} is designed for offline action recognition, in which videos are saved, and then processed offline. To perform recognition for this model in real-time, using Musical Chair, we distribute computations effectively among devices and utilize a sliding window over acquired data.
}

%

\begin{figure}[b]
\centering
\vspace{-10pt}
\includegraphics[width=1.0\linewidth]{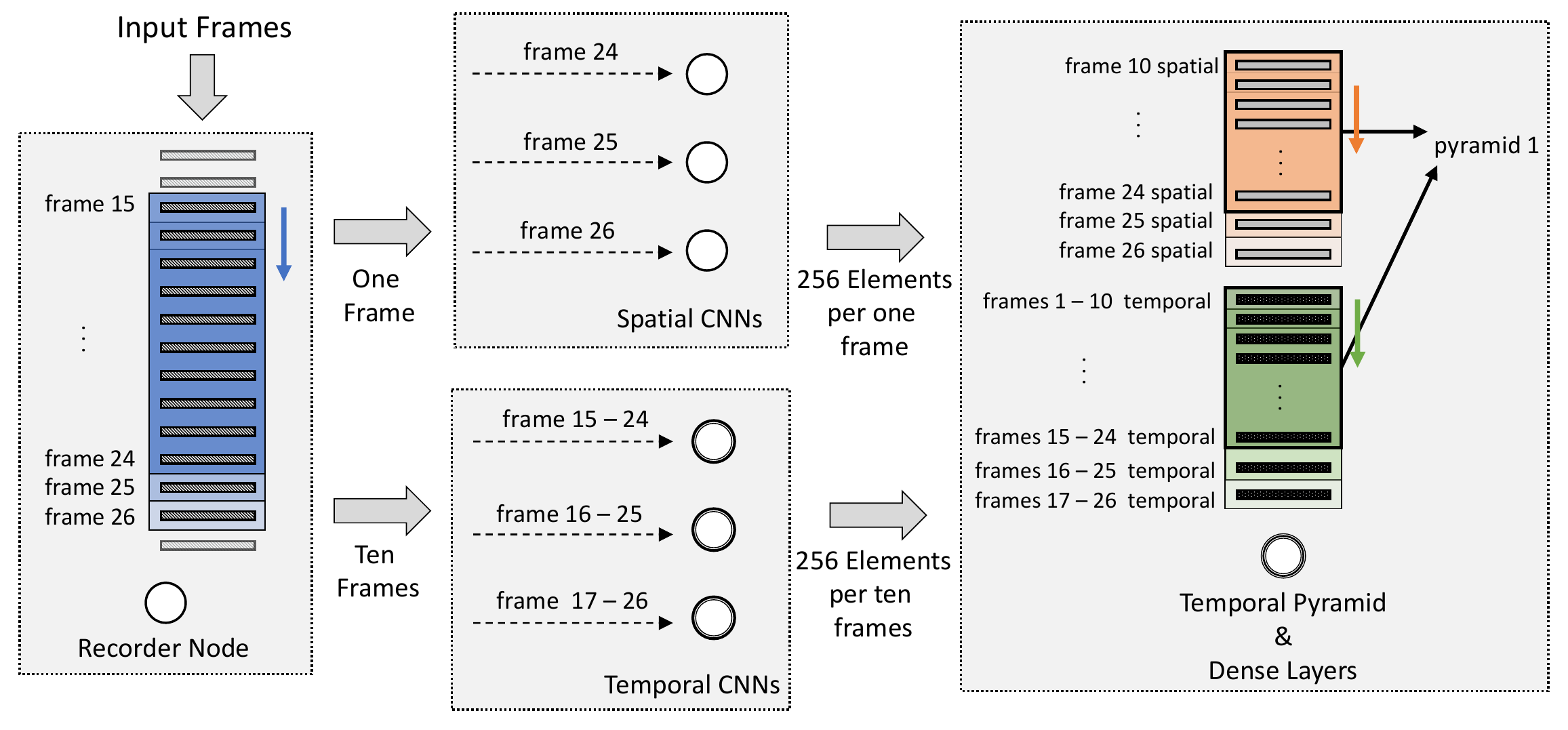}
\captionsetup{singlelinecheck=on,aboveskip=5pt, belowskip=0pt}
\caption{Sliding window for an example system of eight devices. While some tasks require sliding window, with different sizes, others may not need it.}
\label{fig:sliding-window}
\vspace{-0pt}
\end{figure}

\noindent
\textbf{Sliding Window:}
\label{sec:sliding-window}
\noindent
The action recognition model, presented in Section~\ref{sec:two-stream-head}, processes a whole video file for each inference. However, in reality, the frames of a video are generated seamlessly by a camera (30\,FPS). To adapt the model for real-time processing, we propose the use of a sliding window over the input and intermediate data, whenever needed, while distributing the DNN model. For instance, the temporal stream CNN (Section~\ref{sec:temporal}) accepts an input of optical flow from 10 consecutive frames. Therefore, a sliding window of size 10 over the recent inputs is required. In a sliding window, whenever new data arrives, we remove the oldest data and add the new data to the sliding window. Note that to order arriving data, a unique tag is assigned to each raw data during recording time. Figure~\ref{fig:sliding-window} illustrates this point with an example of eight devices in a system. The recorder device keeps a sliding window of size 10 to supply the data for the temporal CNNs, while the spatial and temporal CNNs do not have a sliding window buffer. On the other hand, since the temporal pyramid calculation requires a spatial data of 15 frames and temporal data of 25 frames, the last device keeps two sliding window buffers with different sizes. We can extend the sliding window concept to other DNN models that have a dependency between its inputs to create a continuous data flow. Furthermore, the sliding window is required to enable data and model parallelism. This is because a device needs to order its input data while buffering arrived unordered data.

\noindent
\textbf{Dynamic Task Assignment:}
\noindent
As discussed in Section~\ref{sec:comm}, after a change of condition in the environment, such as detecting motion in other nodes, a game of musical chairs is triggered. The outcome of the game is a device that performs real-time recording, and other devices that process the model. Since all the devices share the same network and they have similar characteristics, the way we assign different tasks to different devices is not important. In other words, we can assign task A to device 1 or 2. However, the loading time of a task is a deciding factor in how long it takes a new system after musical chair game to be ready for real-time processing (i.e., setup time. In our examples setup time is less than a minute). Therefore, when dynamically assigning a new task to a device, Musical Chair considers the previous task of a device to minimize the setup time.

\noindent
\textbf{Supporting Multiple Input Streams:}
\noindent
Musical Chair is able to process multiple input streams. During initialization, the user defines the maximum number of devices ($n_{max}$) . This is mainly because the desirable performance of the system is defined by the total number of available devices. Then, when detecting a motion in the video stream of a device, Musical Chair starts processing the input utilizing more devices to get the best performance. Whenever another input is activated, a new task assignment is performed for creating independent sets of devices for processing each input. Note that, as Section~\ref{sec:music-chair-solution} discussed, the set of tasks for all of the devices is generated in the initialization phase, so no overhead during runtime is associated with supporting multiple input streams or dynamic task assignment.

\section{Evaluation Results}

This section analyzes different system architectures with many Raspberry PIs~\cite{pi3}, the specifications of which are discussed in Section~\ref{sec:nodes}, each with a connected camera~\cite{pi3-cam}. For each system architecture, we provide detailed real-time performance and energy analyses. Although some architectures are restricted by the number of devices and do not necessarily provide the most optimized performance, the insights gained from them are valuable. Furthermore, we compare our results with two implementations: (\romannum{1}) GPU and CPU on a high-performance (HPC) machine (Table~\ref{tab:fu}), and (\romannum{2}) GPU and CPU on a Jetson TX2~\cite{jetson} (Table~\ref{tab:jetson}).

\renewcommand{\arraystretch}{0.85}
\begin{table}[h]
    \small
	\centering
	\vspace{-2pt}
	\captionsetup{singlelinecheck=on,aboveskip=1pt}
	\caption{HPC machine specifications.}
	\begin{tabular}{c | c | c}
		\toprule
        CPU & \multicolumn{2}{c}{2$\x$ 2.00GHz 6-core Intel E5-2620} \\
        Memory &  \multicolumn{2}{c}{1333\,MHz 96\,GB RAM DDR3} \\
        GPU & \multicolumn{2}{c}{Titan Xp (Pascal) 12\,GB GDDR5X} \\
        Total Price & \multicolumn{2}{c}{\$3500} \\
        \midrule
        \multirow{3}{2cm}{\centering Power Consumption }
        &Idle  & 125\,W \\
        &\%100 Only-CPU Utilization & 240\,W \\
        &\%100 Only-GPU Utilization & 250\,W \\
		\bottomrule
	\end{tabular}
	\label{tab:fu}
	\vspace{-10pt}
\end{table} \renewcommand{\arraystretch}{1}

\renewcommand{\arraystretch}{0.85}
\begin{table}[h]
    \small
	\centering
	\vspace{0pt}
	\captionsetup{singlelinecheck=on,aboveskip=1pt}
	\caption{Nvidia Jetson TX2 specifications~\cite{jetson}.}
	\begin{tabular}{c | c | c}
		\toprule
        \multirow{2}{*}{\centering CPU } & \multicolumn{2}{c}{2.00\,GHz Dual Denver 2 +} \\
        & \multicolumn{2}{c}{2.00\,GHz Quad Core ARM Cortex-A57} \\
        \arrayrulecolor{black!20}\midrule\arrayrulecolor{black}
        Memory &  \multicolumn{2}{c}{1600\,MHz 8\,GB RAM LPDDR4} \\
        GPU & \multicolumn{2}{c}{Pascal Architecture - 256 CUDA Core} \\
        Total Price & \multicolumn{2}{c}{\$600} \\
        \midrule
        \multirow{3}{2cm}{\centering Power Consumption }
        &Idle (Power Gated)  & 5\,W \\
        &\%100 Utilization & 15\,W \\
        &Averaged Observed & 9.5\,W \\
		\bottomrule
	\end{tabular}
	\label{tab:jetson}
	\vspace{-5pt}
\end{table} \renewcommand{\arraystretch}{1}

\subsection{Action Recognition}
\label{sec:res-act}


\begin{figure}[t]
\begin{tabular}{c}
\vspace{0pt}

\begin{subfigure}{\columnwidth}
  \includegraphics[width=1.0\linewidth]{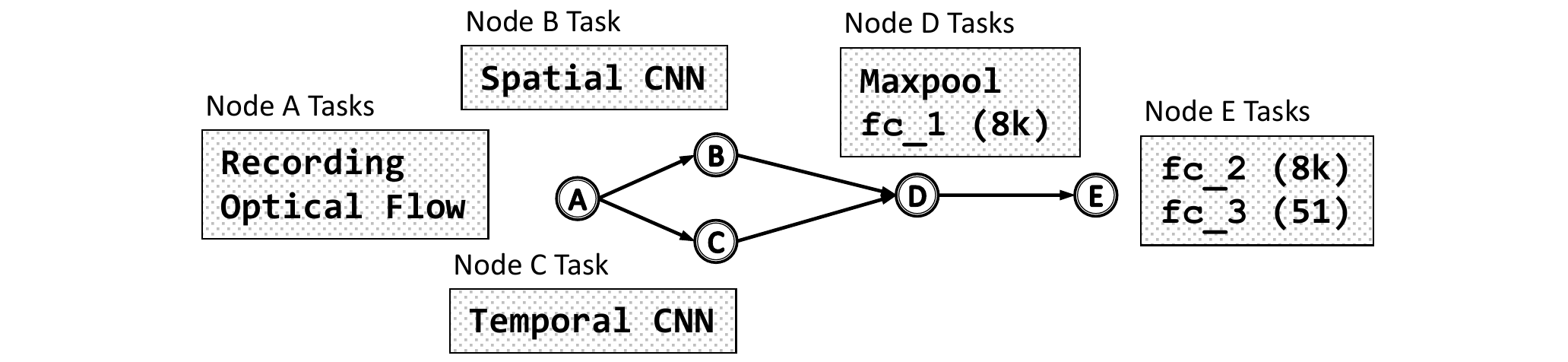}
  \captionsetup{singlelinecheck=on,aboveskip=5pt, belowskip=0pt}
  \caption{Five-device system. Exploiting model parallelism for \texttt{fc} layers.}
  \label{fig:five-node}
\end{subfigure}
\vspace{10pt}

\\

\begin{subfigure}{\columnwidth}
  \includegraphics[width=1.0\linewidth]{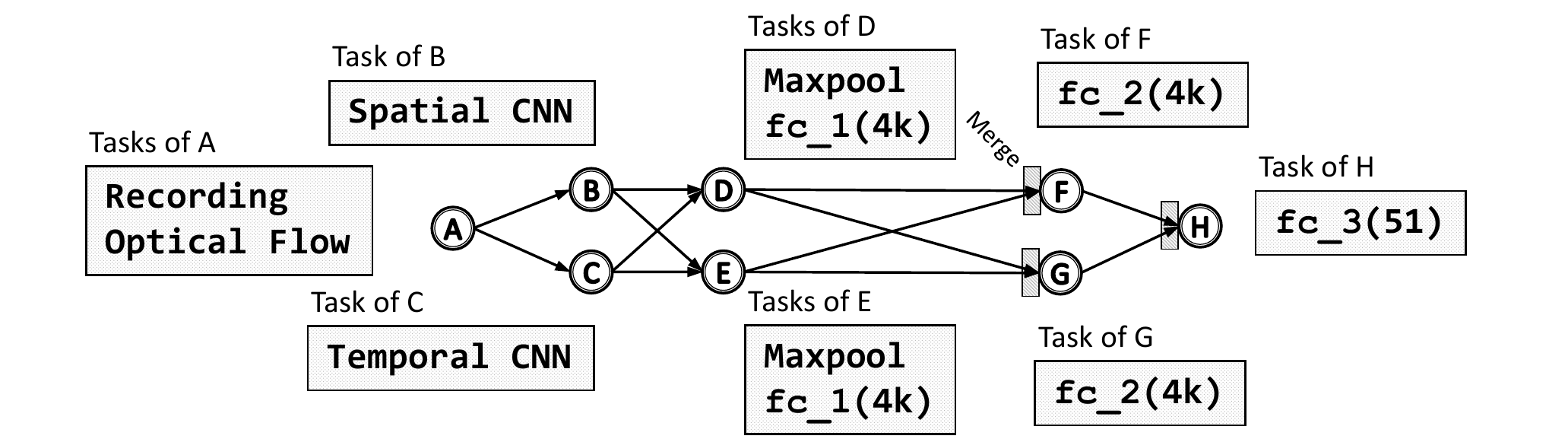}
  \captionsetup{singlelinecheck=on,aboveskip=5pt, belowskip=0pt}
  \caption{Eight-device system. Exploiting model parallelism for \emph{each} \texttt{fc} layer.}
  \label{fig:eight-node}
\end{subfigure}
\vspace{10pt}

\\

\begin{subfigure}{\columnwidth}
  \vspace{0pt}
  \includegraphics[width=1.0\linewidth]{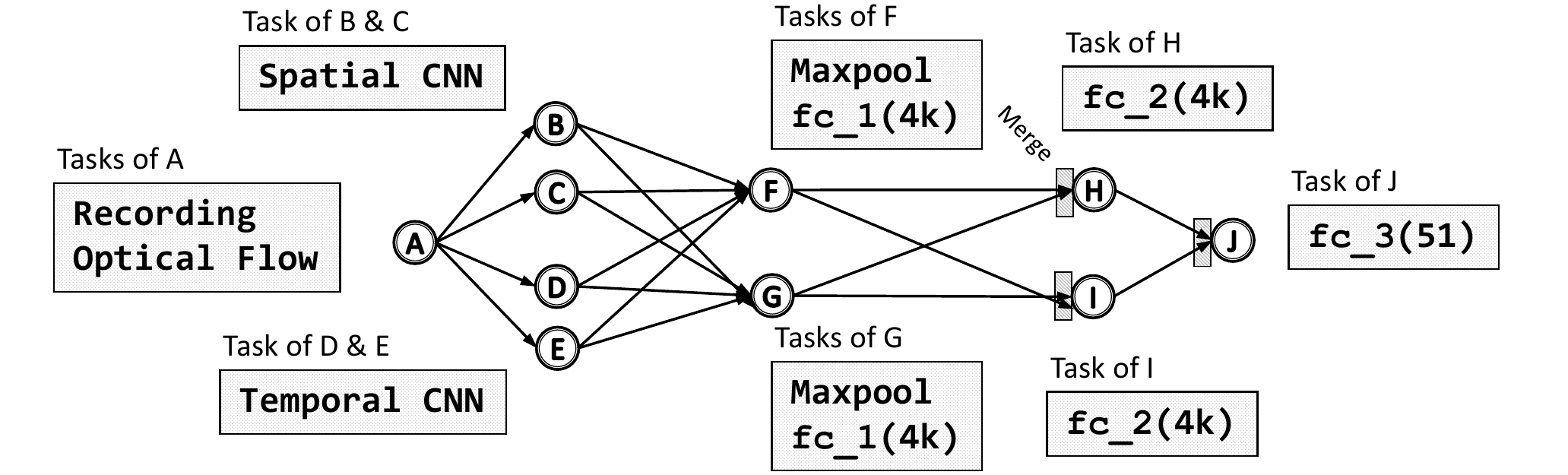}
  \captionsetup{singlelinecheck=on,aboveskip=5pt, belowskip=0pt}
  \caption{10-device system. Exploiting model parallelism for \emph{each} \texttt{fc} layer, and data parallelism for temporal and spatial CNNs.}
  \label{fig:ten-node}
  \vspace{0pt}
\end{subfigure}
\vspace{10pt}

\\

\begin{subfigure}{\columnwidth}
  \includegraphics[width=1.0\linewidth]{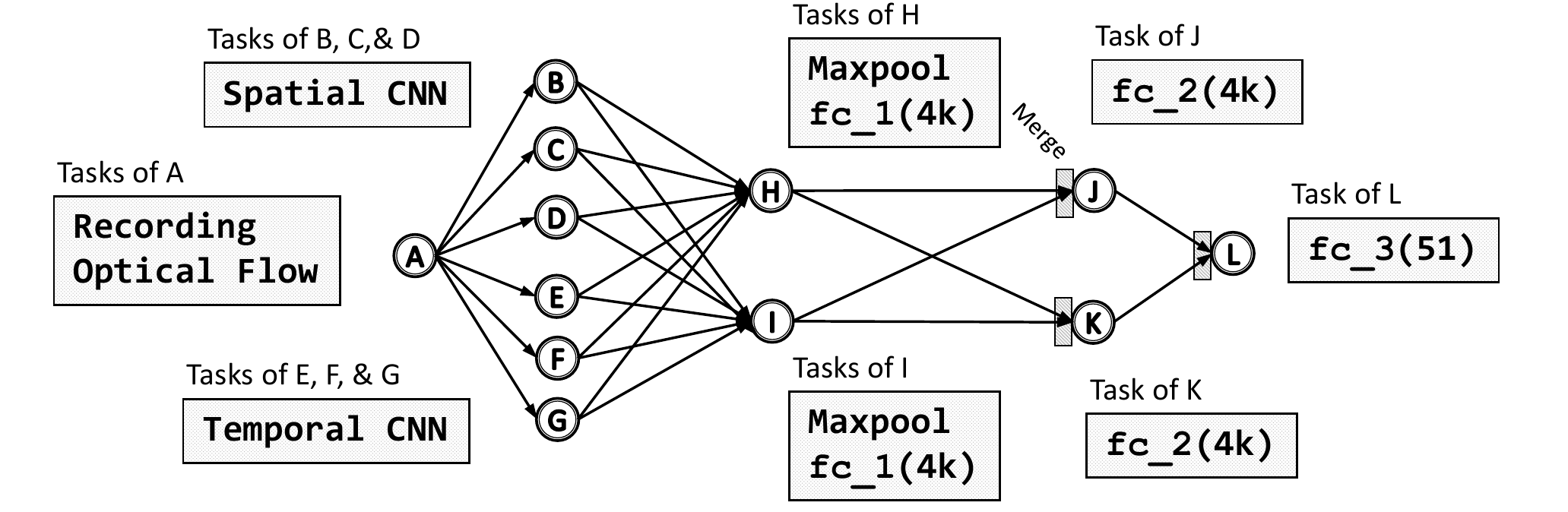}
  \captionsetup{singlelinecheck=on,aboveskip=5pt, belowskip=0pt}
  \caption{12-device system. Exploiting model parallelism for \emph{each} \texttt{fc} layer, and data parallelism for temporal and spatial CNNs.}
  \label{fig:twelve-node}
  \vspace{0pt}
\end{subfigure}

\vspace{10pt}
\end{tabular}

\vspace{-12pt}
\caption{System architectures of action recognition.}  
\label{fig:system-arch}
\vspace{-15pt}
\end{figure}

\noindent
\textbf{System Architectures:}
\noindent
In action recognition, the recording device, which receives camera inputs, also computes optical flow, the computation of which is not heavy (e.g., for 100 frames, 0.004 seconds). Each device manages a sliding window buffer, explained in Section~\ref{sec:sliding-window}, the size of which is dependent on model and data parallelism of previous devices and the input of next device.

\noindent
{\bf \emph{(I) One-Device System:}} For a fair comparison, we
implement a one-device system, in which one device executes all
tasks. As shown in Figure~\ref{fig:nodes-simple}a, in this system, all
tasks are assigned to a single device. Since one device cannot load the full
model in its memory, it needs to load each task sequentially,
generate the intermediate outputs, and then continue to the next
task. Note that, even though our current model is performing recognition
on $12\x 16$ frames, the inference requires a  significant amount of memory. Generally, DNN models utilize a large amount of memory, the provision of which is a challenge for resource-constrained devices. Hence, investigating such memory-limited scenarios provides insights into execution overheads of DNNs.

\noindent
{\bf \emph{(II) Four-Device System (less accurate):}} This system,
depicted in Figure~\ref{fig:nodes-simple}b, has the minimum number of
devices so that reloading the tasks is not necessary. However, to fit
dense layers in one device, we must use the half-sized dense layers
(i.e., 4k-4k-51), which reduces the accuracy by 15\%. Since temporal and spatial CNN are independent, Musical Chair assigns them to devices that work in parallel.

\noindent
{\bf \emph{(III) Five-Device System:}} To use the original \texttt{fc}
layers and reach the maximum accuracy, at least five devices are
required. Musical Chair distributes the final \texttt{fc} layers among
two devices: the first with the 8k \texttt{fc} layer and the other with the 8k and
51 \texttt{fc} layers. This system achieves the same accuracy as that
of the original model.

\noindent
{\bf \emph{(IV) Eight-Device System:}} With more available devices, Musical Chair performs model parallelism on each 8k \texttt{fc} layers, creating two 4k \texttt{fc} layers per each device. Figure~\ref{fig:eight-node} depicts this system. The devices in the same vertical column are executing their tasks in parallel.

\noindent
{\bf \emph{(V) 10-Device System:}} In the 10-device system, two more devices process temporal and spatial CNNs exploiting data parallelism. Therefore, two devices process each CNN stream, illustrated in Figure~\ref{fig:ten-node}. New frames and optical flows are assigned in a round-robin fashion to two devices (of each stream) and are ordered using tags in subsequent devices.

\noindent
{\bf \emph{(VI) 12-Device System:}} With 12 devices, Musical Chair
increases data parallelism for temporal and spatial CNNs by assigning
three devices to each stream, shown in Figure~\ref{fig:twelve-node}.

\begin{figure}[t]
\centering
\vspace{-5pt}
\includegraphics[width=1.0\linewidth]{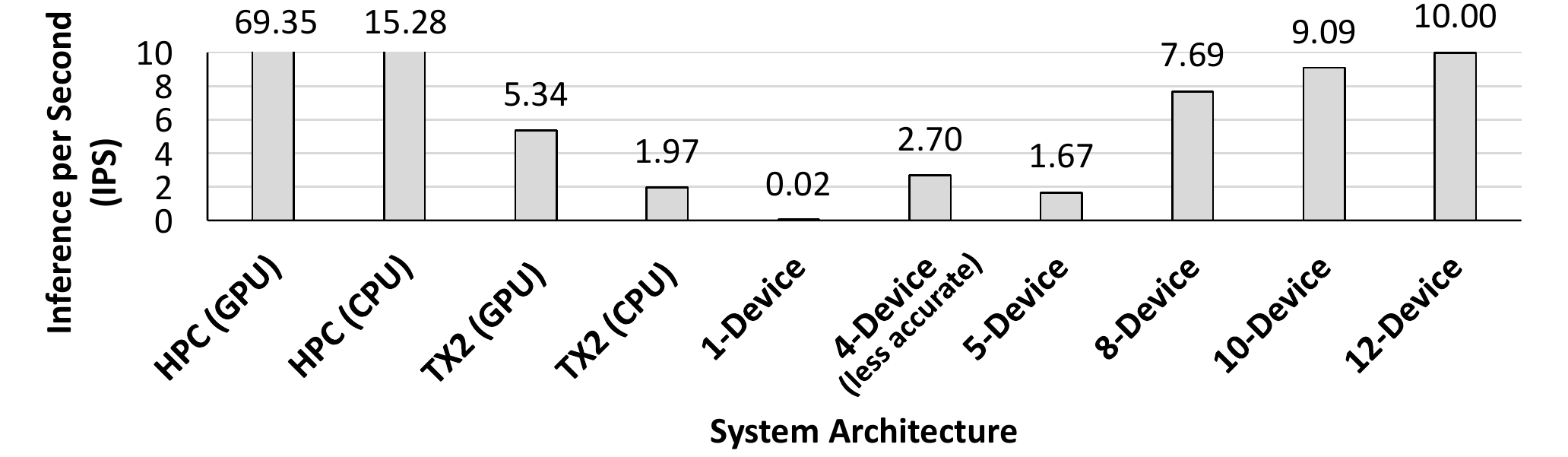}
\captionsetup{singlelinecheck=on,aboveskip=2pt, belowskip=0pt}
\caption{Measured inference per second.}
\label{fig:frame-rate}
\vspace{-18pt}
\end{figure}

\begin{figure}
\centering
\begin{tabular}{c}
\vspace{0pt}

\begin{subfigure}{\columnwidth}
  \centering
  \vspace{0pt}
  \includegraphics[width=1.0\linewidth]{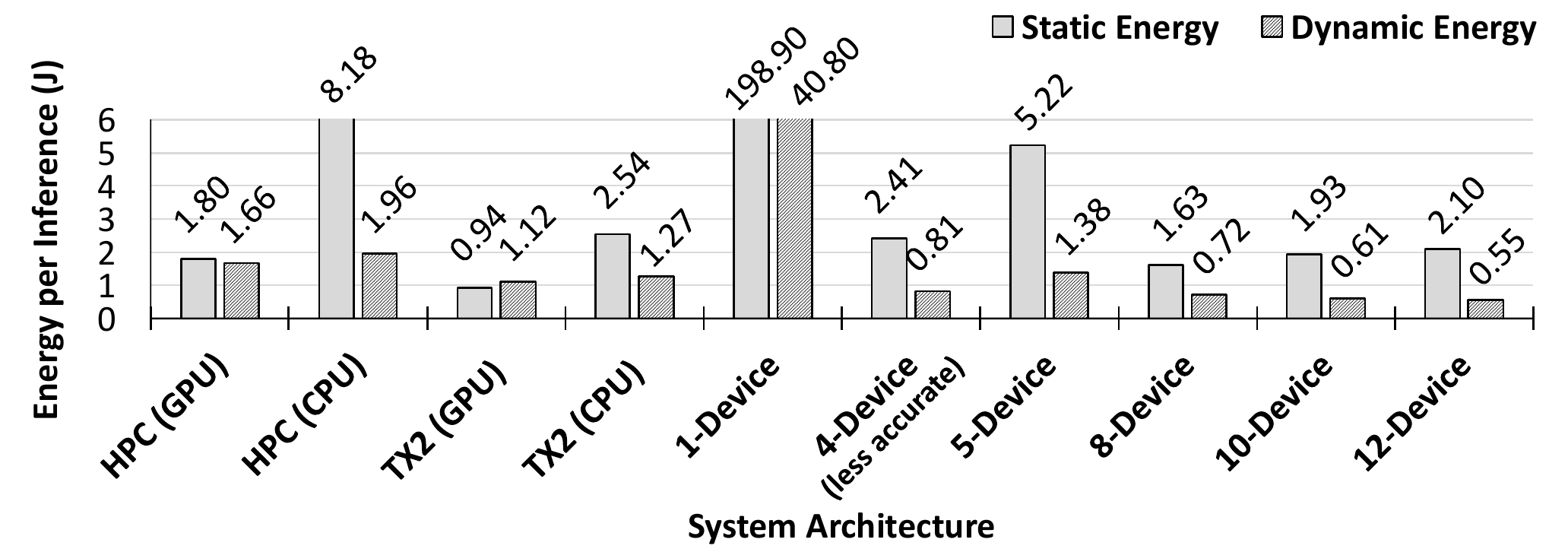}
  \captionsetup{singlelinecheck=on,aboveskip=2pt, belowskip=0pt}
  \caption{Measured static and dynamic energy consumption per inference.}
  \label{fig:energy}
  \vspace{0pt}
\end{subfigure}
\vspace{5pt}

\\

\begin{subfigure}{\columnwidth}
  \centering
  \vspace{0pt}
  \includegraphics[width=1.0\linewidth]{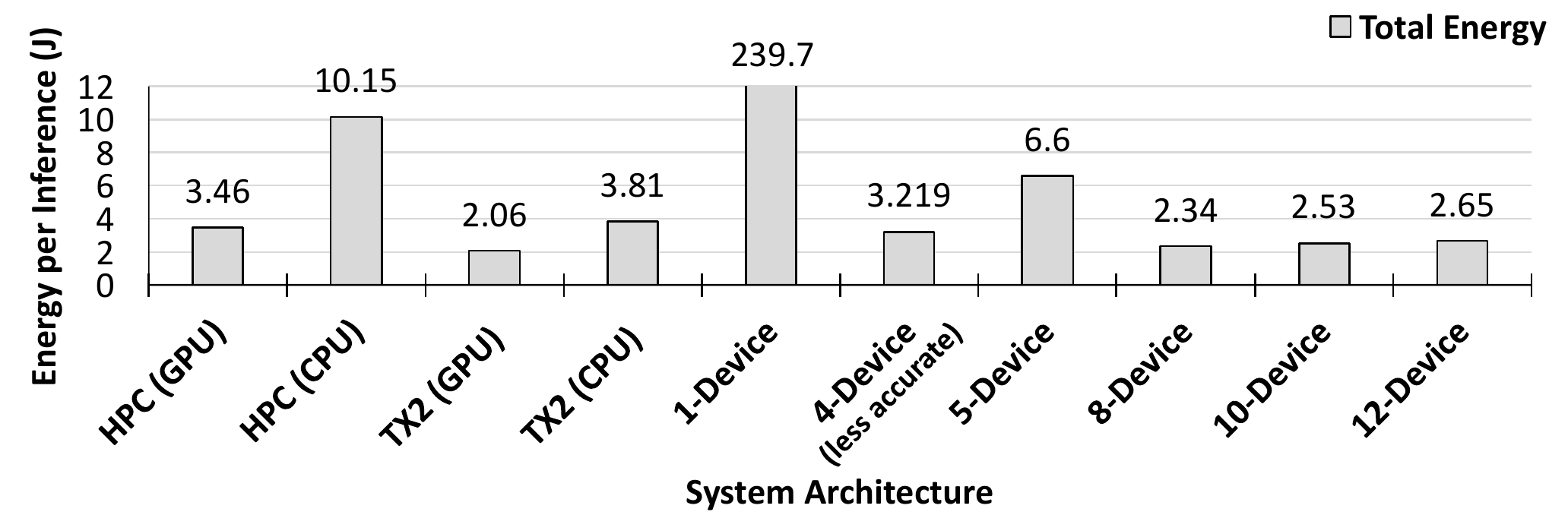}
  \captionsetup{singlelinecheck=on,aboveskip=2pt, belowskip=0pt}
  \caption{Measured total energy consumption per inference.}
  \label{fig:energy-total}
  \vspace{0pt}
\end{subfigure}
\vspace{10pt}

\end{tabular}

\vspace{-14pt}
\caption{Energy consumption per inference.}  
\label{fig:energy-figs-all}
\vspace{-15pt}
\end{figure}

\noindent
\textbf{Performance and Energy Analyses:}
\begin{figure}[b]
\centering
\vspace{-15pt}
\includegraphics[width=1.0\linewidth]{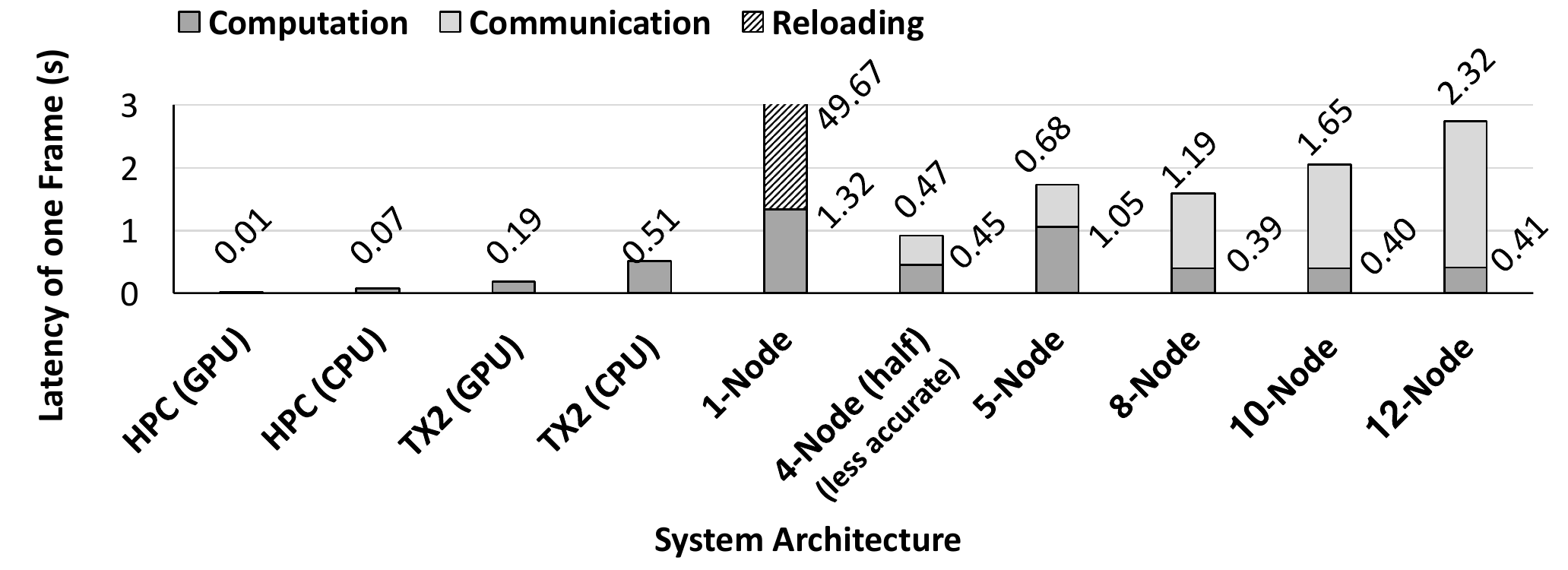}
\captionsetup{singlelinecheck=on,aboveskip=2pt, belowskip=0pt}
\caption{Measured end-to-end latency of one frame.}
\label{fig:latency}
\vspace{-5pt}
\end{figure}
Figure~\ref{fig:frame-rate} presents $IPS$ values for all system architectures, the HPC machine, and the Jetson TX2. As expected, the one-device system performs poorly since one PI needs to compute all the tasks. As we expected, when the number of devices increases, Musical Chair achieves better performance by harnessing the computational power of additional devices. In fact, systems larger than five devices always perform better than the TX2 (speed up of 1.9$\x$ over TX2-GPU and 5$\x$ over TX2-CPU). Figures~\ref{fig:latency}, \ref{fig:energy}, and~\ref{fig:energy-total} display the end-to-end latency of one frame in the system (split in communication, computation, and reloading time), dynamic and static energy consumption per one inference, and total energy consumption per one inference, respectively. Similarly, we observe significant energy consumption and high latency for the one-device system. In other words, in comparison with the one-device system, Musical Chair achieves a 90$\x$ energy savings and a speedup 500$\x$ for $IPS$. As Figure~\ref{fig:latency} illustrates and discussed before, the reason for such speedups is that the one-device system spends most of its latency in reloading subparts of the model.

As Figures~\ref{fig:frame-rate} and \ref{fig:latency} depict, although increasing the number of devices in a system also increases the communication latency notably, we observe a performance gain in the inference per second with a higher number of devices. This is because in both data and model parallelism, the systems hide latency by distributing or parallelizing tasks. In other words, in Figure~\ref{fig:system-arch}, devices in the same vertical column reduce the effective latency for the devices in the next column by parallelizing.
For the large number of devices, Musical Chair achieves not only similar energy consumption with TX2 but also saves energy as compared to the HPC machine. Figure~\ref{fig:energy-total} depicts that, except for the TX2 with GPU, the energy consumptions per inference (i.e., \nicefrac{Watt}{performance}) of systems with more than five devices is always better than other cases (up to 4.3$\x$ and an average of 1.5$\x$). Note that in our evaluations, the power consumption of the Raspberry PI systems is inclined to higher energy consumption because: (\romannum{1}) In comparison with TX2, when we increase the number of devices for Raspberry PI systems, since each device is on a development board and has several unnecessary peripherals, the energy consumption increases significantly, which is shown in static energy; (\romannum{2}) The TX2 is a low-power design with power gating capabilities that power gates three cores if not needed, but the Raspberry PIs do not have such capabilities; and (\romannum{3}) the energy consumption of the Raspberry PI systems also includes the energy for communication between the devices, and wasted energy of powering an idle core during data transmission.\footnote{For Tegra platform, we are aware that the order of our reported numbers are not matching~\cite{nvidia-whitepaper}, but matches another independent work~\cite{can:pas16} }

\subsection{Image Recognition}
\label{sec:res-image}

\begin{figure}[t]
\centering
\vspace{-5pt}
\includegraphics[width=1.0\linewidth]{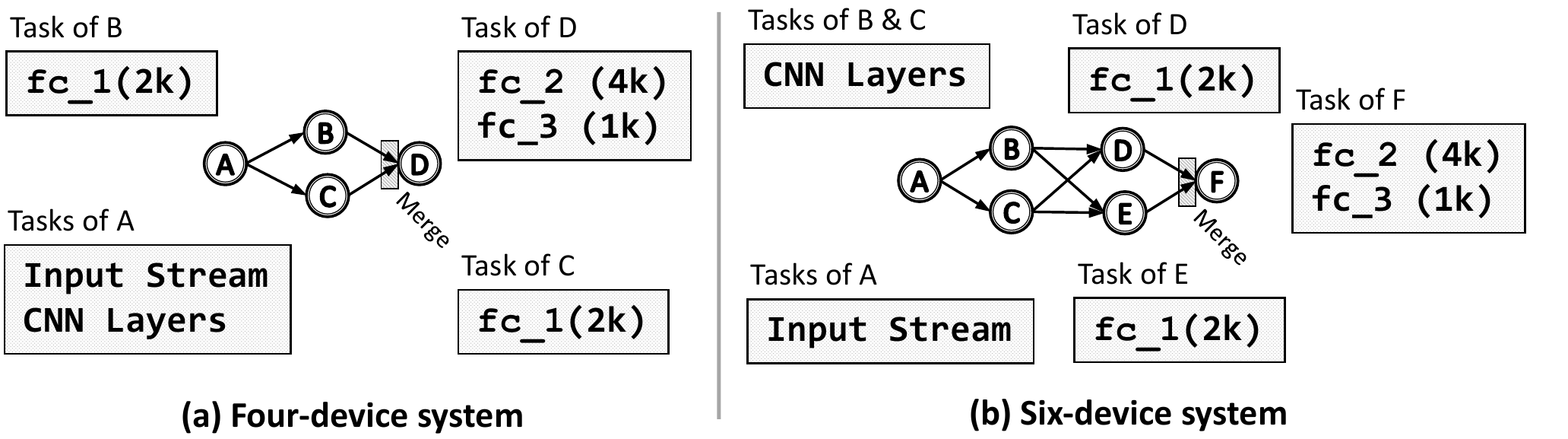}
\captionsetup{singlelinecheck=on,aboveskip=5pt, belowskip=0pt}
\caption{System architectures for AlexNet.}
\label{fig:alexnet-system}
\vspace{0pt}
\end{figure}

\begin{figure}[t]
\centering
\vspace{-5pt}
\includegraphics[width=1.0\linewidth]{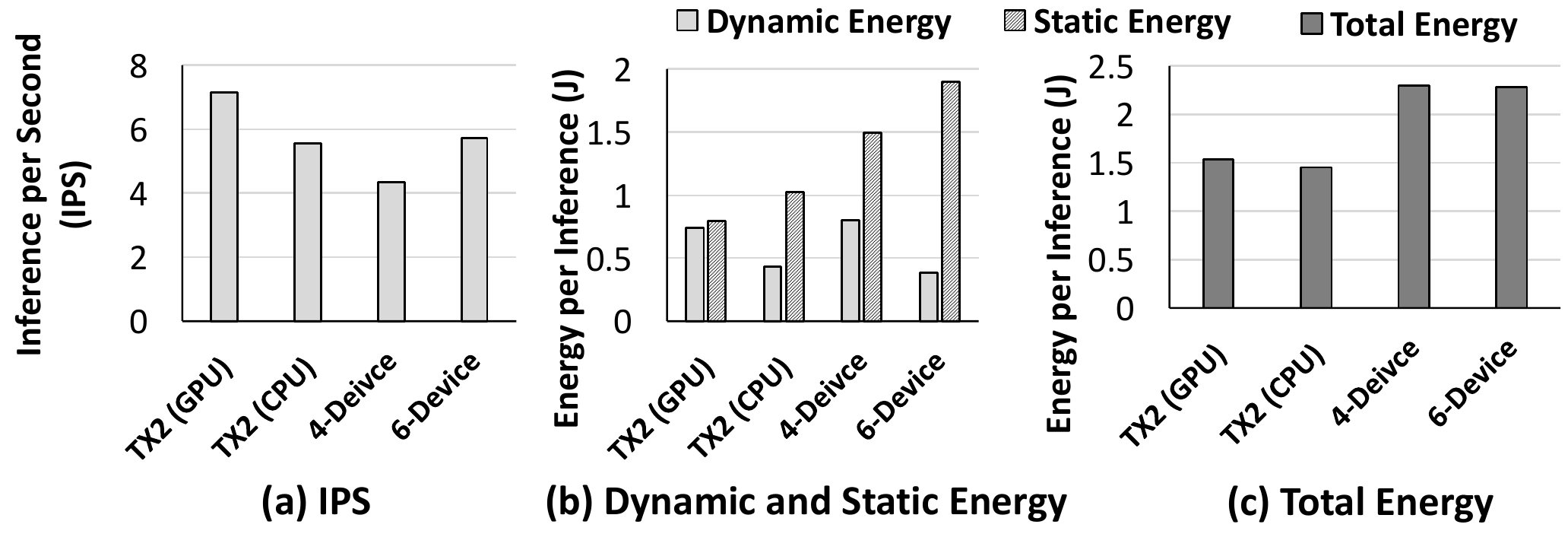}
\captionsetup{singlelinecheck=on,aboveskip=5pt, belowskip=0pt}
\caption{AlexNet: Measured IPS (a), static and dynamic energy consumption (b), and total energy consumption (c).}
\label{fig:alexnet-res}
\vspace{-15pt}
\end{figure}

\noindent
\textbf{AlexNet:}
We apply Musical Chair to AlexNet, the model of which is shown in Figure~\ref{fig:alexnet} that is composed of CNN layers and three \texttt{fc} layers. As a test case, Figures~\ref{fig:alexnet-system}a and b display the generated tasks for the four- and six-device systems, respectively. While in the four-device system, model parallelism is performed on the \texttt{fc\_1} layer, in the six-device system, an additional data parallelism is performed on CNN layers. We implement both systems and measure their performance and energy consumption, shown in Figure~\ref{fig:alexnet-res}. Figure~\ref{fig:alexnet-res}a depicts a performance increment by increasing the number of devices in a system. In fact, the achieved performance of the six-device system is similar to the TX2 with CPU, and 30\% worst than the TX2 with GPU. Furthermore, as discussed in the previous section, Figure~\ref{fig:alexnet-res}b shows that most of the energy consumption of the Raspberry PI systems is because of the static energy consumption.

\begin{figure}[b]
\centering
\vspace{-10pt}
\includegraphics[width=1.0\linewidth]{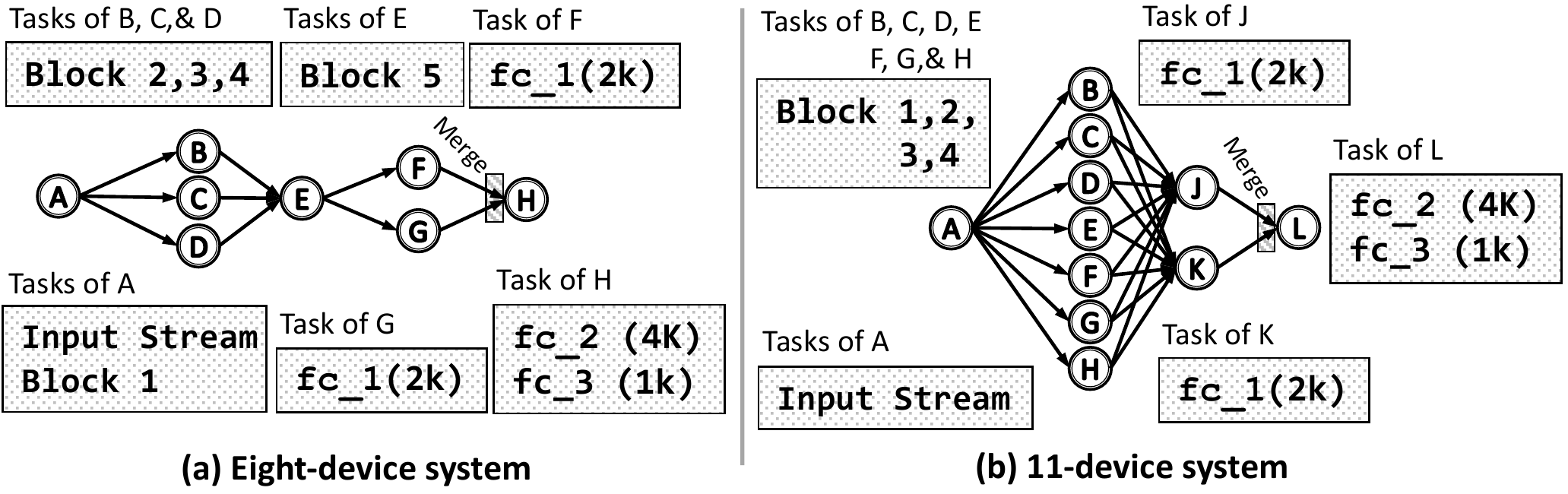}
\captionsetup{singlelinecheck=on,aboveskip=5pt, belowskip=0pt}
\caption{System architectures for VGG16.}
\label{fig:vgg16-system}
\vspace{-0pt}
\end{figure}

\begin{figure}[!b]
\centering
\vspace{-5pt}
\includegraphics[width=1.0\linewidth]{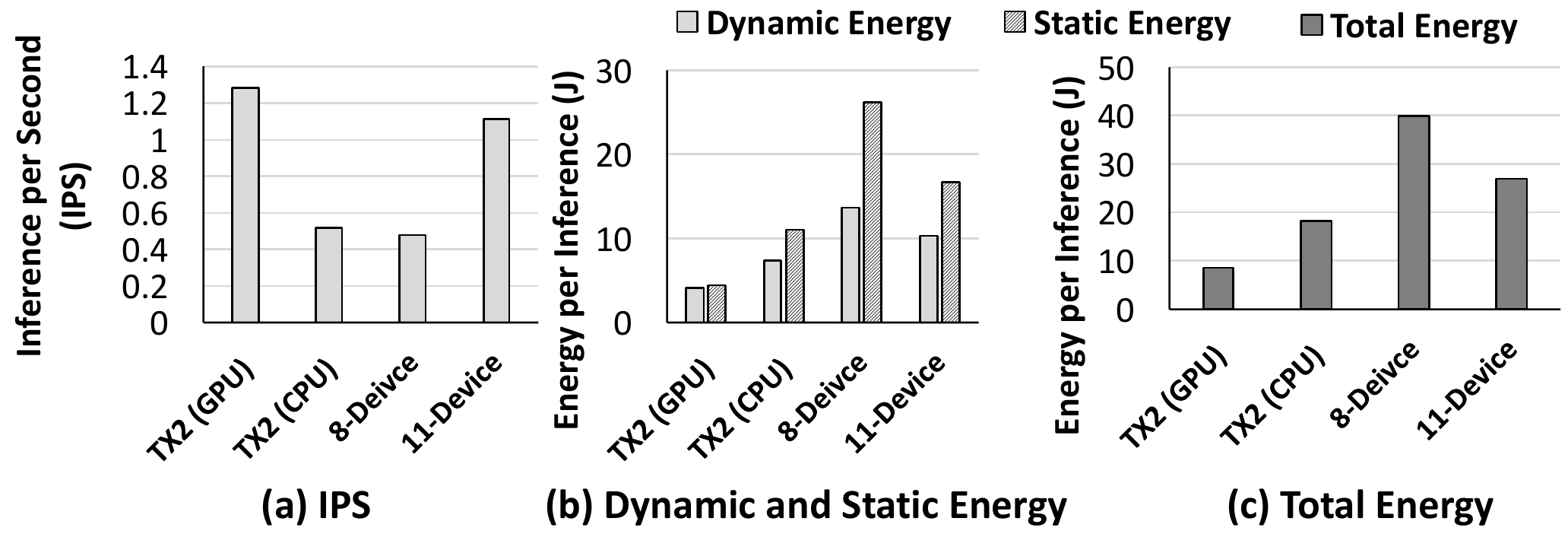}
\captionsetup{singlelinecheck=on,aboveskip=5pt, belowskip=0pt}
\caption{VGG16: Measured IPS (a), static and dynamic energy consumption (b), and total energy consumption (c).}
\label{fig:vgg16-res}
\vspace{-5pt}
\end{figure}

\renewcommand{\arraystretch}{0.85}
\begin{table*}[t]
  \small
  \centering
  \vspace{-5pt}
  \captionsetup{singlelinecheck=on,aboveskip=1pt,belowskip=0pt}
  \caption{Musical Chair comparison with recent related work.}
  \begin{tabular}{l | c | c | c | c | c | c | c}
  \toprule
  & \multirow{2}{1.9cm}{\centering \textbf{End-Compute Device}} 
  & \multirow{2}{1.4cm}{\centering \textbf{Number of Devices}} 
  & \multirow{2}{1.2cm}{\centering \textbf{Localized Inference}}
  & \multirow{2}{1.7cm}{\centering \textbf{Real-Time Data Process}}
  & \multirow{2}{1.8cm}{\centering \textbf{Partitioning Mechanism}}
  & \multirow{2}{2cm}{\centering \textbf{Model- \& Data-Parallelism}}
  & \multirow{2}{1.6cm}{\centering \textbf{Runtime Adaptability}}
  \\
  & & & & & & &\\
  \midrule
  \textbf{Neurosurgeon~\cite{kan:hau17}} 
          & Tegra TK1~\cite{jetson-tk} 
          & 1
          & \xmark
          & \xmark
          & Inter-Layer
          & \xmark
          & \xmark
          \\
  \textbf{MoDNN~\cite{mao:chn17}} 
          & LG Nexus 5
          & 4
          & \cmark
          & \xmark
          & Intra-Layer
          & \xmark
          & \xmark
          \\
  \textbf{DDNN~\cite{tee:mcd17}} 
          & \xmark 
          &  Many  
          & \xmark
          & \cmark
          & Inter-Layer
          & Data Parallelism 
          & \xmark 
          \\
  \textbf{Musical Chair} & Raspberry PI~\cite{pi3}
          & Many
          & \cmark
          & \cmark
          & Intra- \& Inter-Layer
          & Both
          & \cmark
          \\
  \bottomrule
  \end{tabular}
  \label{tab:related}
  \vspace{-13pt}
\end{table*}
\renewcommand{\arraystretch}{1}

\noindent
\textbf{VGG16:}
VGG16 (Figure~\ref{fig:vgg16}), in comparison with AlexNet, is more computationally intensive~\cite{can:pas16}. In order to distribute the model, Musical Chair divides the VGG16 model to several blocks of sequential CNNs. Figures~\ref{fig:vgg16-system}a and~\ref{fig:vgg16-system}b depict the outcome of task assignment for VGG16 with eight and 11 devices, respectively. Note that for \texttt{fc\_1}, since its input size is large, Musical Chair performs model parallelism. On the other hand, for \texttt{fc\_2} and \texttt{fc\_3}, since their computations are not a bottleneck, Musical Chair assigns them to a single device. We measure the performance and energy consumption of both systems and the TX2, shown in Figure~\ref{fig:vgg16-res}. When the number of devices increases from eight to 11, Musical Chair achieves 2.3$\x$ better performance by reassigning all CNN blocks to a device and performing more optimal data parallelism. In fact, compared to the TX2 with GPU, the 11-device system achieves comparable IPS (15\% degradation).

Both the AlexNet and VGG16 results show that Musical Chair provides scalability as the number of devices increases in image recognition models. Since both models are more computationally intensive with large input sizes than the two-stream CNN, a larger number of PIs is required to achieve better performance than the TX2 with GPU. Nonetheless, our system shows similar performance and dynamic energy consumption to the TX2 with GPU with less cost. 

\vspace{-5pt}

\section{Related Work}
Processing DNN models in real-time has various applications in real life. Currently, the processing of DNN models is offloaded to cloud services. A recent work, Neurosurgeon~\cite{kan:hau17}, dynamically partitions a DNN model between a \emph{single} edge device (Tegra TK1, \$200) and the cloud for higher performance and better energy consumption. The partitioning is done based on prediction models and optimizes the amount of transferred data (e.g., in 3G networks) and amount of aggregated computation time in the cloud and edge devices. Neurosurgeon still relies on cloud service availability and does not discuss real-time processing. In addition, all \texttt{fc} layer computations are offloaded to the cloud. Although Neurosurgeon improves the end-to-end latency of computations by 3.1$\x$, it does not provide scalability and fully local real-time DNN processing. A similar study of partitioning between mobile and cloud is done in~\cite{hau:man14} using the Galaxy S3. Another work, MoDNN~\cite{mao:chn17} creates a local distributed mobile computing system and accelerated DNN computations. MoDNN uses only mobile platforms (LG Nexus 5, \$350) and partitions the DNN using input partitioning within each layer especially relying on sparsity in the matrix multiplications. However, MoDNN does not consider real-time processing, as their most optimized system with four Nexus 5 devices MoDNN has a latency of six seconds. Since the partitioning is done within each layer, it incurs high data communication traffic. Furthermore, the number of devices in their experiments does not exceed four. DDNN~\cite{tee:mcd17} also aims to distribute the computation in local devices. However, in its mechanism, in addition to retraining the model, each sensor device performs the first few layers in the network and the rest of the computation is offloaded to the cloud system. Furthermore, it only evaluate proposed algorithms using simulations and confirm accuracy only without proving whether the partitioned network can indeed run on IoT devices. Table~\ref{tab:related} provides a comparison of these works with Musical Chair.

Executing DNN models in resource-constrained platforms has recently gained great attention. For instance, Microsoft created a library (ELL)~\cite{ell} that designs and deploys intelligent machine-learned models onto resource-constrained platforms, such as	Raspberry Pi, Arduino, and micro:bit. Currently, this library includes only image classification models with maximum accuracy of 60\%. The models are tailored for these platforms such that they have a smaller number of weights, and convolution layers have strides of two for reducing the dimensions of the input. Interestingly, these models do not have any fully connected layers since these layers require a large amount of computation. Although such an effort might alleviate the overhead of DNNs on resource-constrained platforms, the lower accuracy of the models in addition to the time consuming exploration of finding a specialized tailored model hinders the implementation of other models, such as action recognition models. Recently, optimizing networks for mobile or embedded platforms has been proposed. As another example, Tensorflow Lite~\cite{tensorflowLite} was just released to support mobile systems.  Although these solutions target mobile platforms, such platforms are still more powerful than embedded systems. In fact, Musical Chair can take advantage of these optimized libraries to reduce the computation and memory requirement for each device. Moreover, the scalability of Musical Chair enables cost-efficient IoT devices to perform complex DNN computations with the highest accuracy achievable by HPC servers.

Another direction is to reduce the overhead of DNNs. This is because high-accuracy DNN models contain a large number of weights and require high computational capability. In short, these models are a natural fit for high-performance computing. Therefore, there has been a significant effort to reduce the overhead of these models such as pruning~\cite{yu:luk17,han:mao15}, resource partitioning~\cite{she:fer16, guo:yin17}, quantization and low-precision inference~\cite{cou:ben14, gon:li14, van:sen11} , binarizing weights~\cite{li:zha16, cou:hub:16,rast:ord16} and simplifying neural network~\cite{ian:mos16, red:kum16}. As an example, XNOR-Net~\cite{rast:ord16} approximated both filters and input to binary values. Such an approximation enables XNOR-Net to perform real-time processing on CPUs (rather than GPUS) while reducing the accuracy by around 8\%. However, besides not targeting mobile platforms, this solution is not scalable and is performed for image classification workloads. In summary, although many studies reduced the overhead of DNNs or distributed DNNs, (\romannum{1}) they did not study cost-efficient IoT devices, (\romannum{2}) they did not examine conditions and methods for real-time processing of DNNs, and (\romannum{3}) they did not design a collaborative system with many devices.

\vspace{-5pt}

\section{Conclusion}
\noindent
In this paper, we proposed Musical Chair to harvest the computational power of resource-constrained and cost-efficient IoT devices by collaboration. Musical Chair is able to adjust to the inherent dynamics of IoT networks and adapt to the availability of IoT devices for optimal performance. Musical Chair does not rely on cloud services, so it preserves the privacy of the users while reducing the load on data centers. We demonstrated the Musical Chair system by examining and implementing a state-of-the art action recognition model and two well-known image recognition models using multiple Raspberry PIs. We created the environment for DNN models to collaborate and measured performance and energy consumption of several system architectures.


\bibliographystyle{ieeetr}
\bibliography{ref}

\begin{thebibliography}{10}

\bibitem{gartner-iot}
I.~Gartner, ``{Gartner Says 6.4 Billion Connected "Things" Will Be in Use in
  2016, Up 30 Percent From 2015}.''
  \url{https://www.gartner.com/newsroom/id/3165317}, 2015.
\newblock [Online; accessed 11/10/17].

\bibitem{gartner-iot-datacenter}
F.~Biscotti, J.~Skorupa, R.~Contu, {\em et~al.}, ``{The Impact of the Internet
  of Things on Data Centers},'' {\em Gartner Research}, vol.~18, 2014.

\bibitem{lee:lee15}
I.~Lee and K.~Lee, ``{The Internet of Things (IoT): Applications, Investments,
  and Challenges for Enterprises},'' {\em Business Horizons}, vol.~58, no.~4,
  pp.~431--440, 2015.

\bibitem{khan:khan12}
R.~Khan, S.~U. Khan, R.~Zaheer, and S.~Khan, ``{Future Internet: The Internet
  of Things Architecture, Possible Applications and Key Challenges},'' in {\em
  International Conference on Frontiers of Information Technology (FIT)},
  pp.~257--260, IEEE, 2012.

\bibitem{kri:sut12}
A.~Krizhevsky, I.~Sutskever, and G.~E. Hinton, ``{Imagenet Classification With
  Deep Convolutional Neural Networks},'' in {\em Advances in Neural Information
  Processing Systems (NIPS)}, pp.~1097--1105, 2012.

\bibitem{col:wes08}
R.~Collobert and J.~Weston, ``{A Unified Architecture for Natural Language
  Processing: Deep Neural Networks with Multitask Learning},'' in {\em
  International Conference on Machine Learning (ICML)}, pp.~160--167, ACM,
  2008.

\bibitem{bah:cho14}
D.~Bahdanau, K.~Cho, and Y.~Bengio, ``{Neural Machine Translation by Jointly
  Learning to Align and Translate},'' in {\em International Conference on
  Learning Representations (ICLR)}, 2015.

\bibitem{ryoo:kim17}
M.~S. Ryoo, K.~Kim, and H.~J. Yang, ``{Extreme Low Resolution Activity
  Recognition with Multi-Siamese Embedding Learning},'' in {\em Conference on
  Artificial Intelligence (AAAI)}, Feb. 2018.

\bibitem{sim:zis14}
K.~Simonyan and A.~Zisserman, ``{Two-Stream Convolutional Networks for Action
  Recognition in Videos},'' in {\em Advances in Neural Information Processing
  Systems (NIPS)}, pp.~568--576, 2014.

\bibitem{wan:li16}
Y.~Wang, H.~Li, and X.~Li, ``{Re-Architecting the On-Chip Memory Sub-System of
  Machine-Learning Accelerator for Embedded Devices},'' in {\em International
  Conference on Computer-Aided Design (ICCAD)}, pp.~1--6, 2016.

\bibitem{kim:par15}
Y.-D. Kim, E.~Park, S.~Yoo, T.~Choi, L.~Yang, and D.~Shin, ``{Compression of
  Deep Convolutional Neural Networks for Fast and Low Power Mobile
  Applications},'' in {\em International Conference on Learning Representations
  (ICLR)}, 2016.

\bibitem{lei:sen13}
X.~Lei, A.~W. Senior, A.~Gruenstein, and J.~Sorensen, ``{Accurate and Compact
  Large Vocabulary Speech Recognition on Mobile Devices},'' in {\em
  Interspeech}, vol.~1, 2013.

\bibitem{mcd:tee17}
B.~McDanel, S.~Teerapittayanon, and H.~Kung, ``{Embedded Binarized Neural
  Networks},'' in {\em International Conference on Embedded Wireless Systems
  and Networks (EWSN)}, pp.~168--173, 2017.

\bibitem{ban:wan17}
S.~Bang, J.~Wang, Z.~Li, C.~Gao, Y.~Kim, Q.~Dong, Y.-P. Chen, L.~Fick, X.~Sun,
  R.~Dreslinski, {\em et~al.}, ``{14.7 A 288$\mu$W Programmable Deep-Learning
  Processor with 270KB On-Chip Weight Storage Using Non-Uniform Memory
  Hierarchy for Mobile Intelligence},'' in {\em International Solid-State
  Circuits Conference (ISSCC)}, pp.~250--251, 2017.

\bibitem{lik:hou16}
R.~LiKamWa, Y.~Hou, J.~Gao, M.~Polansky, and L.~Zhong, ``{RedEye: Analog
  ConvNet Image Sensor Architecture for Continuous Mobile Vision},'' in {\em
  International Symposium on Computer Architecture (ISCA)}, pp.~255--266, 2016.

\bibitem{kan:hau17}
Y.~Kang, J.~Hauswald, C.~Gao, A.~Rovinski, T.~Mudge, J.~Mars, and L.~Tang,
  ``{Neurosurgeon: Collaborative Intelligence Between the Cloud and Mobile
  Edge},'' in {\em International Conference on Architectural Support for
  Programming Languages and Operating Systems (ASPLOS)}, pp.~615--629, 2017.

\bibitem{hau:man14}
J.~Hauswald, T.~Manville, Q.~Zheng, R.~Dreslinski, C.~Chakrabarti, and
  T.~Mudge, ``{A Hybrid Approach to Offloading Mobile Image Classification},''
  in {\em International Conference on Acoustics, Speech and Signal Processing
  (ICASSP)}, pp.~8375--8379, 2014.

\bibitem{tee:mcd17}
S.~Teerapittayanon, B.~McDanel, and H.~Kung, ``{Distributed Deep Neural
  Networks Over the Cloud, the Edge and End Devices},'' in {\em International
  Conference on Distributed Computing Systems (ICDCS)}, pp.~328--339, 2017.

\bibitem{ell}
Microsoft, ``{Embedded Learning Library (ELL)}.''
  \url{https://microsoft.github.io/ELL/}, 2017.
\newblock [Online; accessed 11/10/17].

\bibitem{rast:ord16}
M.~Rastegari, V.~Ordonez, J.~Redmon, and A.~Farhadi, ``{XNOR-Net: Imagenet
  Classification Using Binary Convolutional Neural Networks},'' in {\em
  European Conference on Computer Vision (ECCV)}, pp.~525--542, Springer, 2016.

\bibitem{how:zhu17}
A.~G. Howard, M.~Zhu, B.~Chen, D.~Kalenichenko, W.~Wang, T.~Weyand,
  M.~Andreetto, and H.~Adam, ``{Mobilenets: Efficient Convolutional Neural
  Networks for Mobile Vision Applications},'' {\em arXiv preprint
  arXiv:1704.04861}, 2017.

\bibitem{han:shn16}
S.~Han, H.~Shen, M.~Philipose, S.~Agarwal, A.~Wolman, and A.~Krishnamurthy,
  ``{MCDNN: An Execution Framework for Deep Neural Networks on
  Resource-Constrained Devices},'' in {\em International Conference on Mobile
  Systems, Applications, and Services (MobiSys)}, 2016.

\bibitem{caffe2Go}
Facebook, ``{Caffe2Go: Delivering real-time AI in the palm of your hand}.''
  \url{https://code.facebook.com/posts/196146247499076/delivering-real-time-ai-in-the-palm-of-your-hand/},
  2017.
\newblock [Online; accessed 11/10/17].

\bibitem{tensorflowLite}
Google, ``{Introduction to TensorFlow Lite}.''
  \url{https://www.tensorflow.org/mobile/tflite/}, 2017.
\newblock [Online; accessed 11/10/17].

\bibitem{ian:mos16}
F.~N. Iandola, S.~Han, M.~W. Moskewicz, K.~Ashraf, W.~J. Dally, and K.~Keutzer,
  ``{SqueezeNet: AlexNet-Level Accuracy with 50x Fewer Parameters and <0.5 MB
  Model Size},'' {\em arXiv preprint arXiv:1602.07360}, 2016.

\bibitem{red:kum16}
J.~Redmon, S.~K. Divvala, R.~B. Girshick, and A.~Farhadi, ``{IEEE} conference
  on computer vision and pattern recognition, {(CVPR)},'' June 2016.

\bibitem{he:zha16}
K.~He, X.~Zhang, S.~Ren, and J.~Sun, ``{Deep Residual Learning for Image
  Recognition},'' in {\em Conference on Computer Vision and Pattern Recognition
  (CVPR)}, pp.~770--778, 2016.

\bibitem{sim:zis14-deep}
K.~Simonyan and A.~Zisserman, ``{Very Deep Convolutional Networks for
  Large-Scale Image Recognition},'' in {\em International Conference on
  Learning Representations (ICLR)}, 2015.

\bibitem{sze:liu15}
C.~Szegedy, W.~Liu, Y.~Jia, P.~Sermanet, S.~Reed, D.~Anguelov, D.~Erhan,
  V.~Vanhoucke, and A.~Rabinovich, ``{Going Deeper with Convolutions},'' in
  {\em Conference on Computer Vision and Pattern Recognition (CVPR)}, pp.~1--9,
  2015.

\bibitem{nest}
N.~Labs, ``{Nest Cam Family}.'' \url{https://nest.com/cameras/}, 2017.
\newblock [Online; accessed 11/10/17].

\bibitem{nest-term}
N.~Labs, ``{Nest Thermostats}.'' \url{https://nest.com/thermostats/}, 2017.
\newblock [Online; accessed 11/10/17].

\bibitem{nest-term-breakdown}
iFixit, ``{Nest Learning Thermostat 2nd Generation Teardown}.''
  \url{https://www.ifixit.com/Teardown/Nest+Learning+Thermostat+2nd+Generation+Teardown/13818},
  2017.
\newblock [Online; accessed 11/10/17].

\bibitem{nest-term-breakdown-arm}
T.~Instruments, ``{Texas Instruments AM3703CUS Sitara ARM Cortex A8
  Microprocessor}.'' \url{http://www.ti.com/lit/ds/sprs616f/sprs616f.pdf},
  2017.
\newblock [Online; accessed 11/10/17].

\bibitem{nest-6core}
arsTechnica, ``{Nest Cam IQ is a \$300 Indoor Camera With a 6-Core
  Processor}.''
  \url{https://arstechnica.com/gadgets/2017/05/nest-cam-iq-is-a-300-indoor-camera-with-a-6-core-processor/},
  2017.
\newblock [Online; accessed 11/10/17].

\bibitem{yu:luk17}
J.~Yu, A.~Lukefahr, D.~Palframan, G.~Dasika, R.~Das, and S.~Mahlke, ``{Scalpel:
  Customizing DNN Pruning to the Underlying Hardware Parallelism},'' in {\em
  International Symposium on Computer Architecture (ISCA)}, pp.~548--560, 2017.

\bibitem{han:mao15}
S.~Han, H.~Mao, and W.~J. Dally, ``{Deep Compression: Compressing Deep Neural
  Network with Pruning, Trained Quantization and Huffman Coding},'' in {\em
  International Conference on Learning Representations (ICLR)}, 2016.

\bibitem{she:fer16}
Y.~Shen, M.~Ferdman, and P.~Milder, ``{Maximizing {CNN} Accelerator Efficiency
  Through Resource Partitioning},'' in {\em Proceedings of the 44th Annual
  International Symposium on Computer Architecture}, International Symposium on
  Computer Architecture (ISCA), 2017.

\bibitem{guo:yin17}
J.~Guo, S.~Yin, P.~Ouyang, L.~Liu, and S.~Wei, ``{Bit-Width Based Resource
  Partitioning for CNN Acceleration on FPGA},'' in {\em International Symposium
  on Field-Programmable Custom Computing Machines (FCCM)}, 2017.

\bibitem{cou:ben14}
M.~Courbariaux, Y.~Bengio, and J.-P. David, ``{Training Deep Neural Networks
  with Low Precision Multiplication},'' {\em arXiv preprint arXiv:1412.7024},
  2014.

\bibitem{gon:li14}
Y.~Gong, L.~Liu, M.~Yang, and L.~Bourdev, ``{Compressing Deep Convolutional
  Networks Using Vector Quantization},'' {\em arXiv preprint arXiv:1412.6115},
  2014.

\bibitem{van:sen11}
V.~Vanhoucke, A.~Senior, and M.~Z. Mao, ``{Improving the Speed of Neural
  Networks on CPUs},'' in {\em Proceeding Deep Learning and Unsupervised
  Feature Learning NIPS Workshop}, vol.~1, p.~4, 2011.

\bibitem{li:zha16}
F.~Li, B.~Zhang, and B.~Liu, ``{Ternary Weight Networks},'' {\em arXiv preprint
  arXiv:1605.04711}, 2016.

\bibitem{cou:hub:16}
M.~Courbariaux, I.~Hubara, D.~Soudry, R.~El-Yaniv, and Y.~Bengio, ``{Binarized
  Neural Networks: Training Deep Neural Networks with Weights and Activations
  Constrained to +1 or- 1},'' {\em arXiv preprint arXiv:1602.02830}, 2016.

\bibitem{can:pas16}
A.~Canziani, A.~Paszke, and E.~Culurciello, ``{An Analysis of Deep Neural
  Network Models for Practical Applications},'' {\em arXiv preprint
  arXiv:1605.07678}, 2016.

\bibitem{pi3}
R.~P. Foundation, ``{Raspberry Pi 3}.''
  \url{https://www.raspberrypi.org/products/raspberry-pi-3-model-b/}, 2017.
\newblock [Online; accessed 11/10/17].

\bibitem{pi3-cam}
R.~P. Foundation, ``{Raspberry Pi 3}.''
  \url{https://www.raspberrypi.org/products/camera-module-v2/}, 2017.
\newblock [Online; accessed 11/10/17].

\bibitem{iof:sze15}
S.~Ioffe and C.~Szegedy, ``{Batch Normalization: Accelerating Deep Network
  Training by Reducing Internal Covariate Shift},'' in {\em International
  Conference on Machine Learning (ICML)}, pp.~448--456, 2015.

\bibitem{rus:den15}
O.~Russakovsky, J.~Deng, H.~Su, J.~Krause, S.~Satheesh, S.~Ma, Z.~Huang,
  A.~Karpathy, A.~Khosla, M.~Bernstein, {\em et~al.}, ``{Imagenet Large Scale
  Visual Recognition Challenge},'' {\em International Journal of Computer
  Vision (IJCV)}, vol.~115, no.~3, pp.~211--252, 2015.

\bibitem{zei:fer14}
M.~D. Zeiler and R.~Fergus, ``{Visualizing and Understanding Convolutional
  Networks},'' in {\em European Conference on Computer Vision (ECCV)},
  pp.~818--833, Springer, 2014.

\bibitem{he:zha15}
K.~He, X.~Zhang, S.~Ren, and J.~Sun, ``{Delving Deep into Rectifiers:
  Surpassing Human-Level Performance on Imagenet Classification},'' in {\em
  International Conference on Computer Vision (ICCV)}, pp.~1026--1034, 2015.

\bibitem{jhu:ser07}
H.~Jhuang, T.~Serre, L.~Wolf, and T.~Poggio, ``{A Biologically Inspired System
  for Action Recognition},'' in {\em International Conference on Computer
  Vision (ICCV)}, pp.~1--8, IEEE, 2007.

\bibitem{kar:tod14}
A.~Karpathy, G.~Toderici, S.~Shetty, T.~Leung, R.~Sukthankar, and L.~Fei-Fei,
  ``{Large-scale Video Classification with Convolutional Neural Networks},'' in
  {\em Conference on Computer Vision and Pattern Recognition (CVPR)},
  pp.~1725--1732, 2014.

\bibitem{lap:mar08}
I.~Laptev, M.~Marszalek, C.~Schmid, and B.~Rozenfeld, ``{Learning Realistic
  Human Actions from Movies},'' in {\em Conference on Computer Vision and
  Pattern Recognition (CVPR)}, pp.~1--8, 2008.

\bibitem{wan:sch13}
H.~Wang and C.~Schmid, ``{Action Recognition with Improved Trajectories},'' in
  {\em International Conference on Computer Vision (ICCV)}, pp.~3551--3558,
  2013.

\bibitem{ryo:rot17}
M.~S. Ryoo, B.~Rothrock, C.~Fleming, and H.~J. Yang, ``{Privacy-Preserving
  Human Activity Recognition from Extreme Low Resolution.},'' in {\em
  Conference on Artificial Intelligence (AAAI)}, pp.~4255--4262, 2017.

\bibitem{cho:jeo08}
J.~Choi, W.~J. Jeon, and S.-C. Lee, ``{Spatio-Temporal Pyramid Matching for
  Sports Videos},'' in {\em International Conference on Multimedia Information
  Retrieval (ICMR)}, pp.~291--297, 2008.

\bibitem{lev:gol14}
O.~Levy and Y.~Goldberg, ``{Neural Word Embedding as Implicit Matrix
  Factorization},'' in {\em Advances in Neural Information Processing Systems
  (NIPS)}, pp.~2177--2185, 2014.

\bibitem{far:gun03}
G.~Farneb{\"a}ck, ``{Two-Frame Motion Estimation Based on Polynomial
  Expansion},'' {\em Image Analysis}, pp.~363--370, 2003.

\bibitem{kue:jhu11}
H.~Kuehne, H.~Jhuang, E.~Garrote, T.~Poggio, and T.~Serre, ``{HMDB: A Large
  Video Database for Human Motion Recognition},'' in {\em International
  Conference on Computer Vision (ICCV)}, pp.~2556--2563, 2011.

\bibitem{siamese}
J.~Bromley, I.~Guyon, Y.~LeCun, E.~S{\"a}ckinger, and R.~Shah, ``{Signature
  Verification Using a Siamese Time Delay Neural Network},'' in {\em Advances
  in Neural Information Processing Systems (NIPS)}, pp.~737--744, 1994.

\bibitem{had:cho06}
R.~Hadsell, S.~Chopra, and Y.~LeCun, ``{Dimensionality Reduction by Learning an
  Invariant Mapping},'' in {\em Conference on Computer Vision and Pattern
  Recognition (CVPR)}, vol.~2, pp.~1735--1742, 2006.

\bibitem{bel:bal15}
S.~Bell and K.~Bala, ``{Learning Visual Similarity for Product Design with
  Convolutional Neural Networks},'' {\em Transactions on Graphics (TOG)},
  vol.~34, no.~4, p.~98, 2015.

\bibitem{coa:huv13}
A.~Coates, B.~Huval, T.~Wang, D.~Wu, B.~Catanzaro, and N.~Andrew, ``{Deep
  Learning with COTS HPC Systems},'' in {\em International Conference on
  Machine Learning}, pp.~1337--1345, 2013.

\bibitem{a53}
A.~holdings, ``{Embedded Learning Library (ELL)}.''
  \url{https://developer.arm.com/products/processors/cortex-a/cortex-a53},
  2017.
\newblock [Online; accessed 11/10/17].

\bibitem{chollet2015keras}
F.~Chollet {\em et~al.}, ``{Keras}.'' \url{https://github.com/fchollet/keras},
  2015.

\bibitem{tensorflow2015-whitepaper}
M.~Abadi {\em et~al.}, ``{{TensorFlow}: Large-Scale Machine Learning on
  Heterogeneous Systems},'' 2015.
\newblock Software available from tensorflow.org.

\bibitem{apache}
T.~A.~S. Foundation, ``{Apache Avro}.'' \url{https://avro.apache.org}, 2017.
\newblock [Online; accessed 11/10/17].

\bibitem{jetson}
NVIDIA, ``{NVIDIA Jetson TX}.''
  \url{http://www.nvidia.com/object/embedded-systems-dev-kits-modules.html},
  2017.
\newblock [Online; accessed 11/10/17].

\bibitem{nvidia-whitepaper}
Nvidia, ``{GPU-Based Deep Learning Inference: A Performance and Power
  Analysis},'' {\em NVidia Whitepaper}, 2015.

\bibitem{jetson-tk}
NVIDIA, ``{NVIDIA TK}.''
  \url{http://www.nvidia.com/object/jetson-tk1-embedded-dev-kit.html}, 2017.
\newblock [Online; accessed 11/10/17].

\bibitem{mao:chn17}
J.~Mao, X.~Chen, K.~W. Nixon, C.~Krieger, and Y.~Chen, ``{MoDNN: Local
  Distributed Mobile Computing System for Deep Neural Network},'' in {\em
  Design, Automation \& Test in Europe Conference \& Exhibition (DATE)},
  pp.~1396--1401, IEEE, 2017.

\end{thebibliography}

\end{document}